%% file: main.tex
\documentclass[letterpaper, 10 pt, conference]{ieeeconf}  %

\IEEEoverridecommandlockouts                              %

\usepackage{graphics} %
\usepackage{epsfig} %
\usepackage{mathptmx} %
\usepackage{times} %
\usepackage{amsmath} %
\usepackage{amssymb}  %
\usepackage[ruled]{algorithm2e}
\usepackage{xcolor}
\usepackage[most]{tcolorbox}
\usepackage{lipsum}
\usepackage{hyperref}
\usepackage[capitalise]{cleveref}
\usepackage{longtable}
\usepackage{subfig}

\newcommand{\shortened}{\textcolor{gray}{\texttt{[... ommitted for space]}}}

\DeclareMathAlphabet{\mathcal}{OMS}{cmsy}{m}{n}

\title{\LARGE \bf
Traffic Scenario Orchestration from Language\\
via Constraint Satisfaction
}

\author{Frieda Rong$^{1}$, Chris Zhang$^{1,2}$, Kelvin Wong$^{1,2}$, and Raquel Urtasun$^{1,2}$%
\thanks{Work done while at Waabi.}%
\thanks{F.R. acknowledges support from an NSERC CGS-D fellowship.}%
\thanks{$^{1}$University of Toronto $^{2}$Waabi}%
}

\begin{document}

\maketitle
\thispagestyle{empty}
\pagestyle{empty}

\input{abstract.tex}

\input{intro.tex}

\input{related.tex}

\input{method.tex}

\input{experiments.tex}

\input{conclusion.tex}
\input{appendix.tex}

\bibliographystyle{IEEEtran}
\bibliography{IEEEabrv,main}

\input{appendix_extended.tex}

\end{document}

%% file: abstract.tex
\begin{abstract}
    Autonomous vehicles (AVs) require extensive testing in simulation, but test case
    generation for driving scenarios is laborious. 
    The desired scenarios are often out-of-distribution and have precise 
    requirements on interactions with the AV policy under test.
    Manually programming scenarios allows for 
    precise controllability but is difficult to scale.
    On the other hand, statistical models can leverage compute and data,
    but struggle with precise controllability when out-of-distribution.
    We cast scenario orchestration as a constraint-solving problem
    and present a language-in, simulation-out scenario orchestrator 
    for closed-loop testing AVs.
    Our approach leverages foundation model reasoning to translate 
    general, natural language descriptions into a set of constraints as a scenario representation.
    This then allows us to leverage off the shelf solvers to solve for actor behaviors 
    which meet precise testing intentions in closed-loop.
    Under a benchmark of carefully crafted and diverse scenario descriptions, 
    our approach greatly outperforms our baselines in orchestration success rate.
    We further show that our closed-loop approach is especially important for
    scenarios which require ego-reactive specifications.
\end{abstract}

%% file: intro.tex
\section{INTRODUCTION}
Simulation for testing has become an industry standard for autonomous vehicle
(AV) development.
By creating scenarios in virtual environments,
simulation offers a safe, efficient and precise way to test and verify AV behavior.
A key aspect of simulation is the \emph{scenario orchestrator},
the mechanism by which the simulation engine places and controls agents in the scene,
in order to fulfill some user specified testing goal.

Orchestration is a difficult and multifaceted task. Consider
the simple goal of testing the AV's reaction to a cut-in with
some particular time-to-collision (TTC).
First and foremost, the orchestrator must be \emph{controllable},
in that there must exist an interface for
the user to be able to specify the test intention.
In this example, the intention involves an \emph{interaction} between an actor
and the AV, so the orchestrator must also be \emph{reactive} to the AV's behavior
in closed-loop. The orchestrator must reason on a high-level when
determining initial conditions of the actors in order to ensure
the test is feasible (e.g. traveling in the same direction, adjacent lanes, etc.).
It must also ensure that the low-level trajectories are
realistic and kinematically feasible. Finally, this is only one possible test
intention; the orchestrator should be generalizable to support a wide diversity
of potential testing scenarios, many of which may be out-of-distribution.

One approach towards meeting the aforementioned requirements
is to specify scenarios using a domain-specific language (DSL), such as
OpenScenario 2.0 \cite{openscenario2}. Then, typically programmatic behavior models
consisting of template maneuvers, triggers, etc. can be composed in
order to control actors to meet the scenario intent.
The main challenge with this structured and programmatic approach is
its rigidity and lack of scalability. Maintaining a scenario
framework or DSL and updating abstractions to meet new demands from
the messy, unstructured real world is costly, and this approach scales neither with
more data nor compute.

\begin{figure}[t]
    \centering
    \includegraphics*[width=\linewidth]{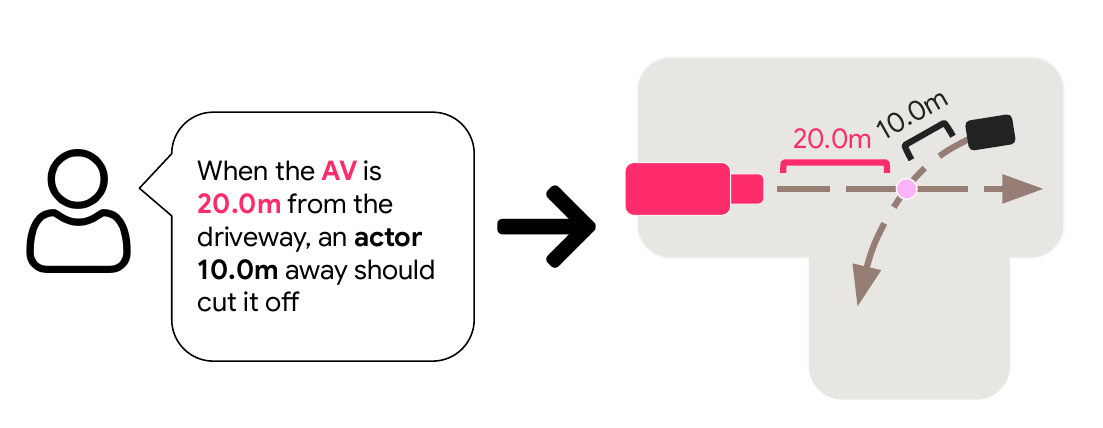}
    \caption{We present a language-in, simulation-out framework for testing autonomous vehicles. We
        leverage LLMs in conjunction with SMT solvers to orchestrate scenarios that meet
        precise user test intentions.}
\end{figure}
\begin{figure*}[t]
    \centering
    \includegraphics[width=\textwidth,trim={0 3.5cm 0 3.5cm},clip]{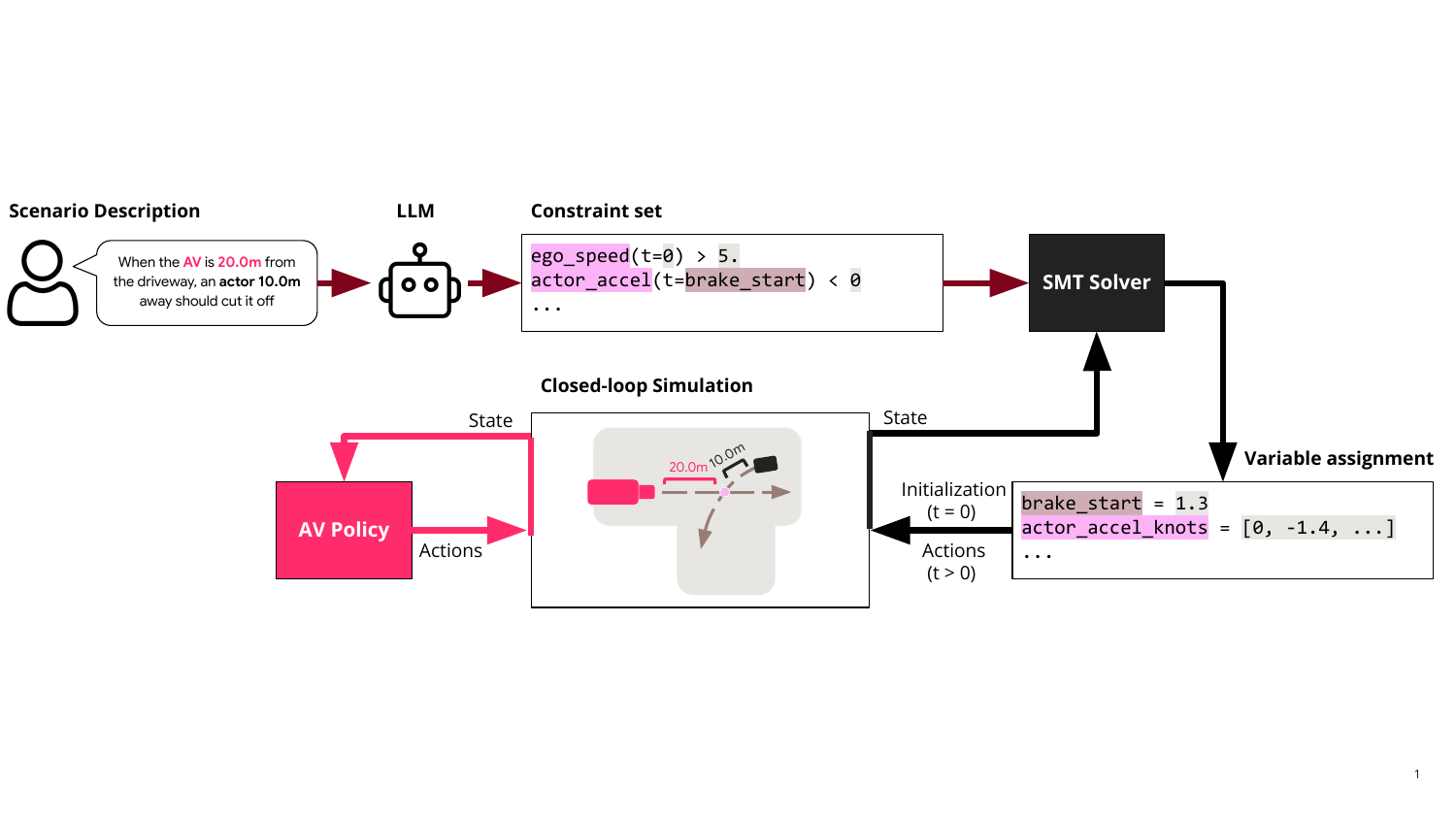}
    \caption{Our overall pipeline. A user inputs a scenario description in
    natural language, which gets converted by an LLM into declarative
    constraints along with actor routes and symbolic motion profiles (not
    shown). The SMT solver runs in closed loop against the uncontrolled AV
    policy to search for satisfying assignments to the constraints which
    determine the concrete actor motion profiles that the simulation framework
    uses to simulate the scenario. 
    }
    \label{fig:overview}
\end{figure*}
Alternatively, recent work opt for more unstructured and learning based
solutions towards orchestration. For instance, \cite{tan2023language,tan2024promptable}
learns a natural language conditioned deep learning based initialization or actor model.
While flexible, these approaches can struggle to orchestrate complex
scenarios that require more reasoning, e.g. \cite{tan2024promptable} cannot support
more complex agent interactions like cut-in or overtake. More generally, we
find that leaving reasoning entirely to an LLM with no tools or assistance can be inefficient and unreliable,
as today's frontier models are still prone to making simple mistakes.
Furthermore, the interface between the LLM and actor model can be overly lossy or coarse.
For instance, specifying coarse spatial bins or relying on latent representations
can make more precise requirements difficult to fulfill.
Finally, there is an inherent tension where data driven methods are trained to
produce in-distribution scenarios, but testing scenarios are often designed
to be edge-case, or contain out-of-distribution commands.

In this work, we propose a \emph{neurosymbolic} approach that combines
the frontier commonsense
reasoning capabilities of LLMs with the precise and controllable reasoning of
classical artificial intelligence algorithms. In particular, we use LLMs in
conjunction with Satisfiability Modulo Theory (SMT) solvers to orchestrate scenarios.
The LLM is tasked to translate natural language descriptions to a set of
constraints, where an off-the-shelf solver can be used to determine the free variable assignments.

Our approach has several advantageous properties.
Firstly, the use of a powerful solver
relieves the LLM of the need to reason about fine-grained
interactions and tedious calculations while applying their strengths in
high-level spatial and temporal reasoning. Conversely, we take advantage of the
low-level functionality of SMT solvers while avoiding the more open-ended
difficulty of formulating the constraint programming problem by using LLMs to
produce the problem formulation.
Our approach efficiently models actors as piece-wise polynomials over space and time,
allowing for a joint modelling of both actor initialization and behaviors.
It also allows for smooth and physically realizable trajectories,
and for constraints to be specified in continuous time and space with arbitrary levels of precision.
Finally, our approach is naturally reactive in closed-loop, as the SMT
solver can easily be re-run at an arbitrary re-planning frequency to take into account
new AV actions.

We evaluate our approach on a set of 80 diverse and challenging natural language scenario descriptions.
Our experiments highlight our approach's efficacy for controllable, precise, and reactive scenario orchestration.
Compared to a state-of-the-art learning-based baseline, our approach achieves a higher orchestration success rate
across a range of scenario families, especially those involving multiple interactions.
We also demonstrate the importance of our closed-loop approach for orchestrating scenarios
with precise interactions between the ego and a hero actor.

%% file: related.tex
\section{RELATED WORK}

\subsection{Programmatic Scenario Generation}
In the self-driving industry, simulation scenarios are
commonly created by programming in domain specific languages
\cite{openscenario2,queiroz2019geoscenario,fremont2019scenic,chang2022metascenario}.
These are typically frameworks that have been designed with composable abstractions
and primitives meant to aid in scenario description or generation.
For instance, OpenScenario 2.0~\cite{openscenario2} provides actions like
\texttt{vehicle.drive()} that can be used to program more complex scenarios.
Scenic~\cite{fremont2019scenic} is a probabilistic programming language,
allowing for programs to define a generative model over scenarios
through sampling individual scenario parameters.
Different approaches support a mixture of imperative or declarative descriptions of scenarios,
and vary in how generation is implemented. In some cases, the language is meant to be purely
descriptive, and generation methodology is left to the user~\cite{openscenario2}. In
others, a rejection sampling approach is taken, generating scenes at random from the imperative descriptions until
all declarative constraints are met \cite{fremont2019scenic}.
Closest to our approach are methods which use solvers to find parameter assignments
which satisfy all declarative constraints~\cite{klischat2020synthesizing,nonnengart2020crisgen,scheibler2019solving,guo2024sovar}.

We aim to address several challenges with programmatic scenario generation faces.
While these frameworks offer controllability for expert designers, manually
writing program specification is difficult to scale. Our work leverages advances in
LLMs to support an auto-formalization of scenarios from natural language descriptions.
Our constraint approach provides an efficient way to generate scenarios without relying
on rejection sampling.
Yet unlike existing constraint based approaches which
are open-loop in nature \cite{klischat2020synthesizing,guo2024sovar}, our method takes a closed-loop
approach where agents adapt to the actions of an external uncontrolled ego vehicle.

\subsection{LLM-Based Scenario Generation}

The survey paper \cite{gao2025foundation} discusses the use of foundation models
in autonomous driving for scenario generation and analysis. They identify
LLM-based scenario generation to be a notable area for research and review
related works. They distinguish between open- and closed-loop scenario
generation, where the latter enables interaction with and reactivity to the ego
vehicle. In our work, we present a solution to LLM-based scenario generation for
both modes of execution. We now highlight a few papers on prior work.

The recent works \cite{cai2025text2scenario,gao2025laser} define a similar
problem of text to scenario generation. The work \cite{gao2025laser} in
particular targets ``on-demand'' scenario execution, allowing a user to program
the system and use LLMs to decompose the user request into scripts and
subscripts for individual rule-based actors to execute at each step of
simulation.
Compared to \cite{gao2025laser}, we invoke the LLM only in the translation stage
and not for decision-making during the scenario execution. Moreover we target
and measure precise adherence to reactive behavior, which neither
\cite{cai2025text2scenario} (which is not reactive) nor \cite{gao2025laser}
does.

LeGEND \cite{tang2024legend} presents an LLM-based prompt engineering
workflow for converting abstract and functional scenarios described in natural
language into logical and concrete scenarios expressed in a structured format,
while ScenicNL \cite{elmaaroufi2024generating} converts crash reports in
natural language into programs written in a custom DSL using a sophisticated
prompting strategy. \cite{nguyen2024texttodrive} converts scenario descriptions
into state charts. These works
are bottlenecked by their DSL or structured
format. %
They do not consider closed-loop reactivity either.

LCTGen \cite{tan2023language} and ProSim \cite{tan2024promptable} are
language-conditional data-driven methods for open- and closed-loop scenario
generation, respectively. In comparison to our approach, these methods are based
on deep learning models and, in the case of ProSim, offer a statistical prior
that can be conditioned on input prompts or other sources of conditioning (e.g.
actions, waypoints or route goals). While the neural network learns to
reconstruct the data, it is only approximately sensitive to conditional input
and may ignore the instructions. Moreover, their closed-loop approach does not
model complex interactions between agents.

\begin{figure}[t]
    \centering
    \includegraphics[width=0.24\linewidth]{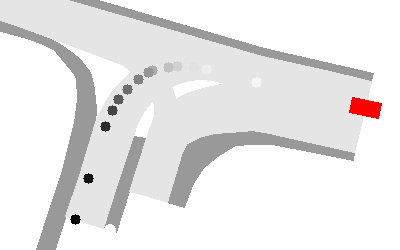}
    \includegraphics[width=0.24\linewidth]{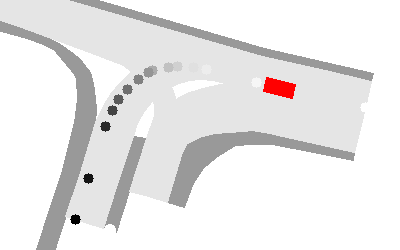}
    \includegraphics[width=0.24\linewidth]{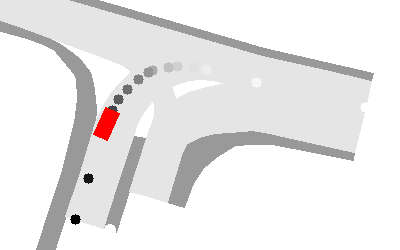}
    \includegraphics[width=0.24\linewidth]{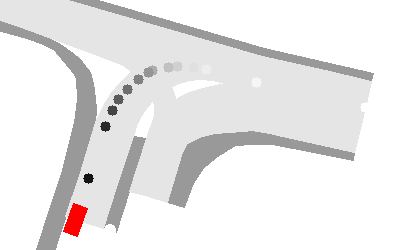}
    \caption{While the solver reasons in 1-D by modelling the kinematics with a piecewise polynomial in Frenet frame,
        the resulting motion profile is projected back into the driving path.}
    \label{fig:turn}
\end{figure}

\subsection{Automated Motion Profiles}
An alternative line of work stemming from the fields of computer graphics and
animation uses formal specifications to generate solutions for  motion profiles.
This idea dates back to the 1980s in the classic ``Spacetime Constraints''
\cite{witkin1988spacetime}. In this paradigm, the user specifies the ``how'' and
``what'' of the motion, while a solver produces a motion satisfying that input
specification. Beyond self-driving applications like \cite{klischat2020synthesizing},
our orchestration approach also shares similarities to the animation system
in \cite{ellman2003automated} in which a series of rewrite rules converts an
initial specification into a lower-level representation to be consumed by an
optimization program or solver based upon kinematic equations of motion. Our
rewrite rules are specific to our polynomial formulation and we present the
details of our term expansion rules in \cref{alg:symbolic_time,alg:concrete_time}.

\noindent
\begin{minipage}[!t]{\linewidth}
  \begin{tcolorbox}[title={\textbf{\scriptsize Sample Natural Language Description of a Scenario}}, colback=white, colframe=black, fontupper=\scriptsize]
    Ego is on the road and it is initial\_ego\_distance\_behind\_intersection\_m behind the intersection. It wants to lane follow along the road travelling westbound. There is a hero vehicle on the road. It wants to turn left into the driveway. When the ego is ego\_distance\_behind\_conflict\_point\_m behind the the conflict point and the hero vehicle is hero\_distance\_behind\_conflict\_point\_m behind the conflict point, the hero vehicle begins to decelerate to a complete stop at end\_hero\_distance\_behind\_conflict\_point\_m behind the conflict point.
  \end{tcolorbox}
  \vspace{5pt}
  \label{box:description}
  Box 1: A sample description fed to the LLM. We parameterize the scenario with
  variables (indicated in snake case) to be assigned values later in the pipeline.
\end{minipage}
\vspace{5pt}

\subsection{Constraint-based Scenario Autoformalization}

The approach we take is a combination of the above and shares similarities with
\cite{guo2024sovar}. The idea is to represent a scenario as a constraint-based
specification and search for a solution to that specification. An LLM can aid by
autoformalizing a natural language description, such as an accident report, into
the expected specification format. The solution is typically actor trajectories.
In \cite{guo2024sovar}, however, the scenario is not closed loop in nature but
rather fixed.

%% file: method.tex
\section{METHODOLOGY}

We propose a neurosymbolic approach to scenario orchestration that combines
an LLM with an SMT solver to orchestrate scenarios from natural language.
Given a scenario description and a high definition (HD) map, we use an LLM
to identify the actors in the scenario and the routes they drive along.
The LLM also specifies each actor's motion profile along its route and
a corresponding set of constraints that govern its kinematics and interactions
with other actors and the map.
Next, an SMT solver solves for the free variables in each actor's motion profile,
determining its initial state and how it moves along its route.
Finally, to orchestrate a closed-loop scenario, we iteratively reinvoke the solver
with additional constraints on all actors' current state to replan each actor's motion profile.
See \cref{fig:overview} for an illustration.

\subsection{Problem Formulation}\label{sec:problem-formulation}
A scenario trace $\mathcal{S}$ consists of an HD map $\mathcal{M}$
and $N$ actor states $\{s_i^{0:T}\}_{i=1}^{N}$ over a finite time horizon $T$.
The $i$-th actor's state $s_i^t$ at time $t$ consists of its 2D position $(x_i^t,y_i^t)$,
heading $\theta_i^t$, velocity $v_i^t$, and acceleration $a_i^t$.
The HD map $\mathcal{M}$ is represented by a lane graph $G = (V, E)$.
Each vertex $u \in V$ is a lane segment and is annotated with various properties;
e.g., whether it is a road or driveway, whether it is a left or right turn, etc.
The edges $(u,v) \in E$ indicate that that $v$ is a successor, predecessor, or left/right neighbor of $u$.

Given a scenario description and an HD map, the goal of \emph{scenario orchestration}
is to generate a scenario trace $\mathcal{S}$ that satisfies the description.
The description declares what should occur; see \hyperref[box:description]{Box 1} for an example.
While it may be given in natural language, we assume that there exists
an underlying function $I\colon\mathcal{S}\mapsto\{0,1\}$ that can verify whether
the trace $\mathcal{S}$ satisfies the description.
Therefore, the orchestrator's goal is to generate $\mathcal{S}$ such that $I(S) = 1$.

In this work, we assume an HD map is given and focus on generating the
actor states $\{s_i^{0:T}\}_{i=1}^{N}$.
Note that the orchestrator may not have control over every actor in the scene.
In particular, the ego vehicle is typically controlled by the policy under test.
The orchestrator must therefore anticipate and react to the ego's behavior to generate a valid trace.

\subsection{Scenario Representation}
\label{sec:scenario-representation}
We represent an actor's trajectory by its motion profile along its driving path;
ie, in Frenet coordinates (\cref{fig:turn}).
We compute the actor's path by traversing the lane graph along its route
and fitting a cubic spline to the centerlines of the lane segments traversed.
From the lane graph, we derive locations of interest along the path;
e.g., when the path enters/exits a left (resp., right) turn lane segment, the intersection, etc.
We also identify any conflict points where two actors' paths intersect.

We model an actor's motion profile with a piecewise polynomial function that represents
the actor's position, velocity, and acceleration along its path over time.
For simplicity, we assume there are no lateral deviations.
The function can have an arbitrary number of pieces $ K $, and each piece $k$ is defined
by its time duration $t_k$ and a polynomial function $ \ell_k $ given by the kinematic equations of motion
\begin{align}
    \ell_0(t) &= x_0 \\
    \ell_k(t) &= \ell_{k-1}(t_{k - 1}) + v_k(t - t_{k - 1}) + \tfrac{1}{2}a_k(t - t_{k - 1})^2
\end{align}
where $x_0$ is the actor's initial position along its path, $v_k$ is its velocity for piece $k$,
$a_k$ is its acceleration for piece $k$, $t_k$ is the time duration of the piece, and $t$ ranges
from $t_1 + \ldots + t_{k - 1}$ to $t_1 + \ldots + t_k$ for each piece $k$.
Thus, an actor's motion is parameterized by the set of variables $ \tau = \{x_0, v_k, a_k, t_k\}_{k = 1}^{K} $
and $\mathcal{T} = \{\tau_i\}_{i = 1}^{N}$ is the joint parameterization for all actors.

We use the Frenet transformation for the actor's path to convert between
Cartesian and Frenet coordinates.
Let $\Gamma_\rho$ be the Frenet transformation for a path $\rho$.
For an actor's state $(x^t, y^t, \theta^t, v^t, a^t)$ at time $t$, we have
\begin{align}
    \ell(t), \dot{\ell}(t), \ddot{\ell}(t) &= \Gamma_\rho(x^t, y^t, \theta^t, v^t, a^t) \\
    (x^t, y^t, \theta^t, v^t, a^t) &= \Gamma_\rho^{-1}(\ell(t), \dot{\ell}(t), \ddot{\ell}(t))
\end{align}

\begin{figure*}[!t]
\begin{minipage}{0.48\linewidth}
\centering
\includegraphics[width=\textwidth]{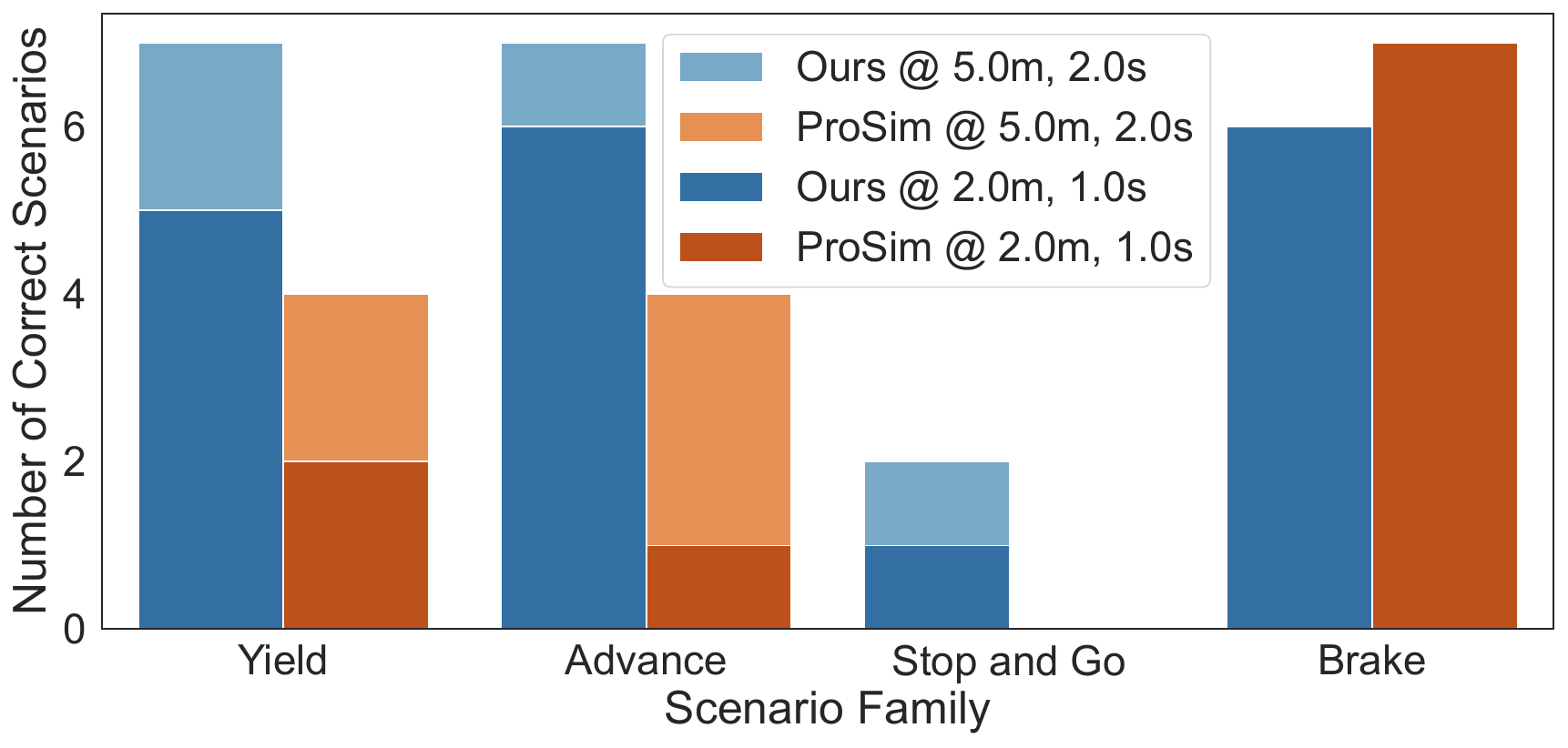}
\captionof{figure}{\textbf{Quantitative Comparison.} More saturated in color and
taller bars indicate better performance. Notice that our method achieves
challenging stop-and-go scenarios.}
\label{fig:quantitative}
\end{minipage}
\hfill
\begin{minipage}{0.48\linewidth}
\centering
\begin{picture}(25,25)
  \put(0,10){\rotatebox{90}{\textsf{\scriptsize ProSim}}}
\end{picture}
\includegraphics[height=5em,trim={8cm 16cm 20cm 8cm},clip]{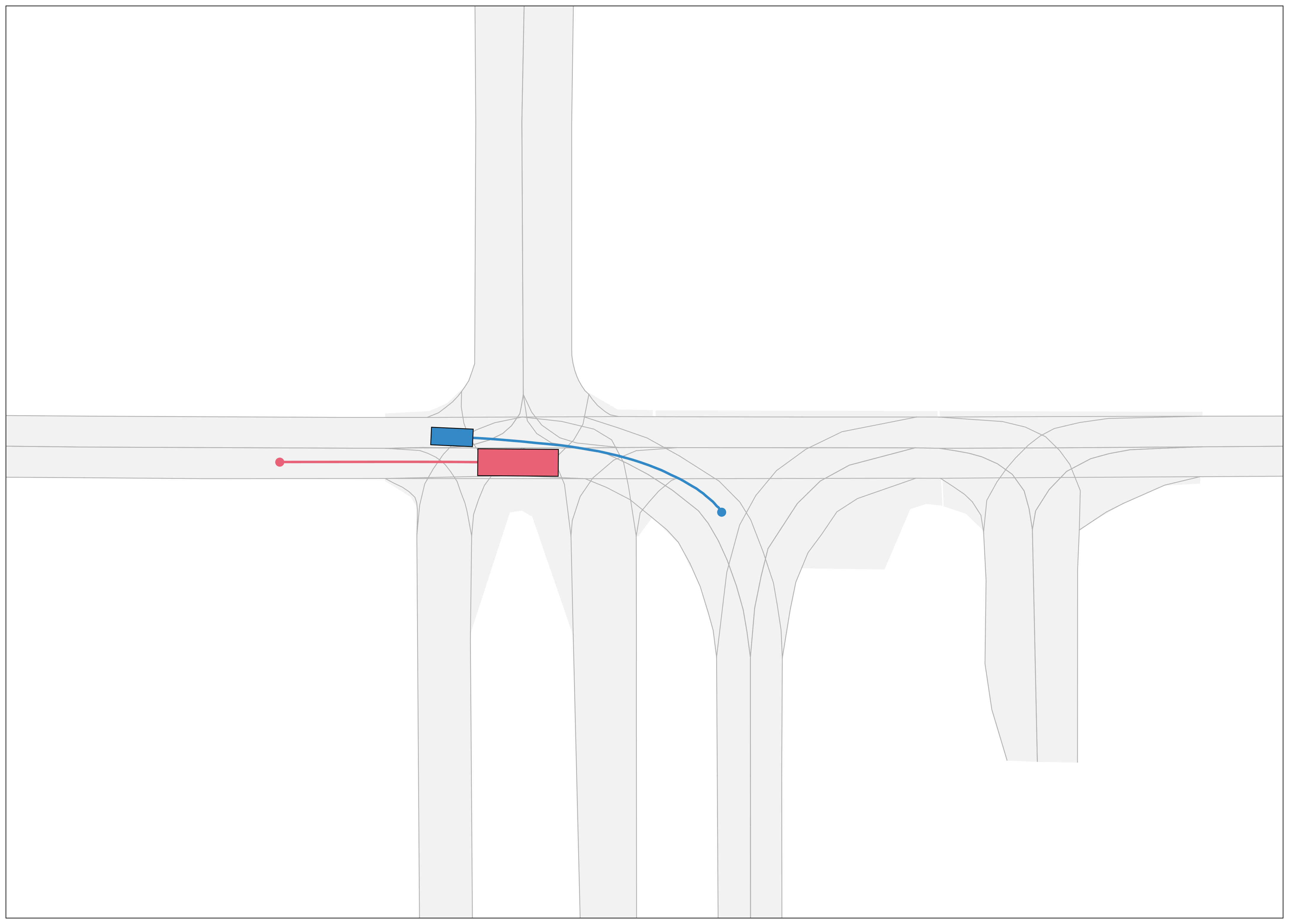}\hspace{2em}
\includegraphics[height=5em,trim={8cm 16cm 20cm 8cm},clip]{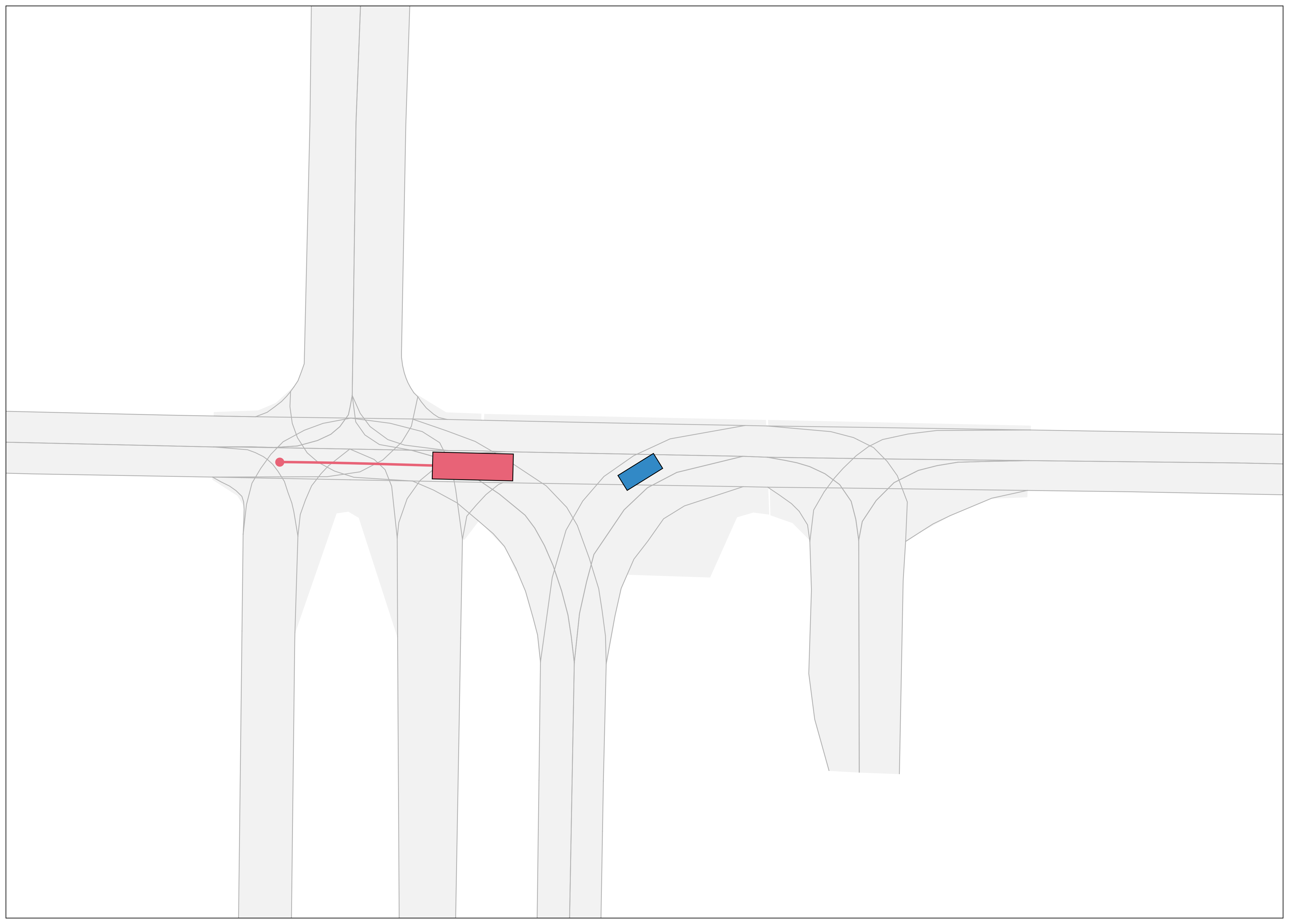}\\\vspace{1em}
\begin{picture}(25,25)
  \put(0,10){\rotatebox{90}{\textsf{\scriptsize Ours}}}
\end{picture}
\includegraphics[height=5em,trim={8cm 16cm 20cm 8cm},clip]{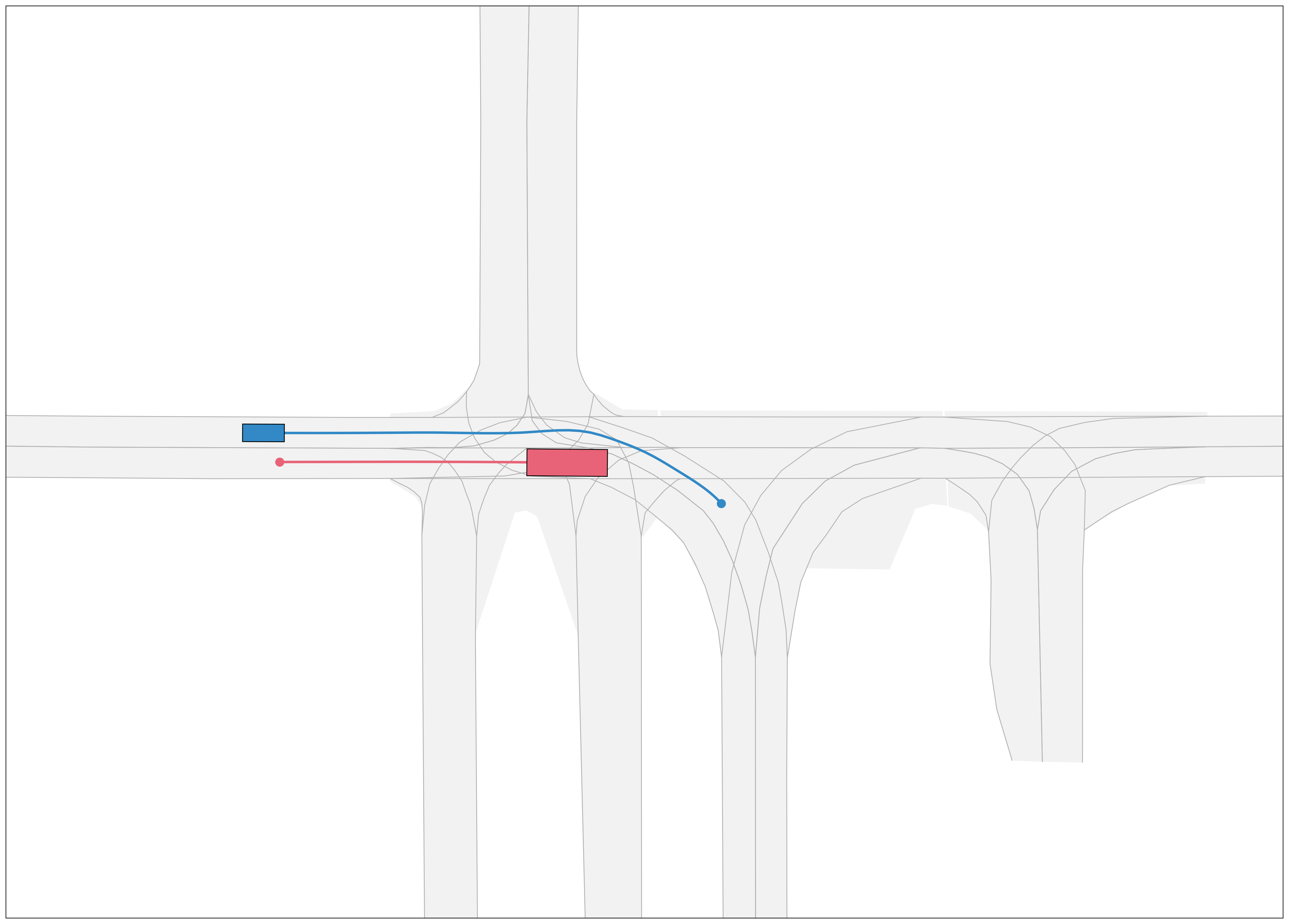}\hspace{2em}
\includegraphics[height=5em,trim={8cm 16cm 20cm 8cm},clip]{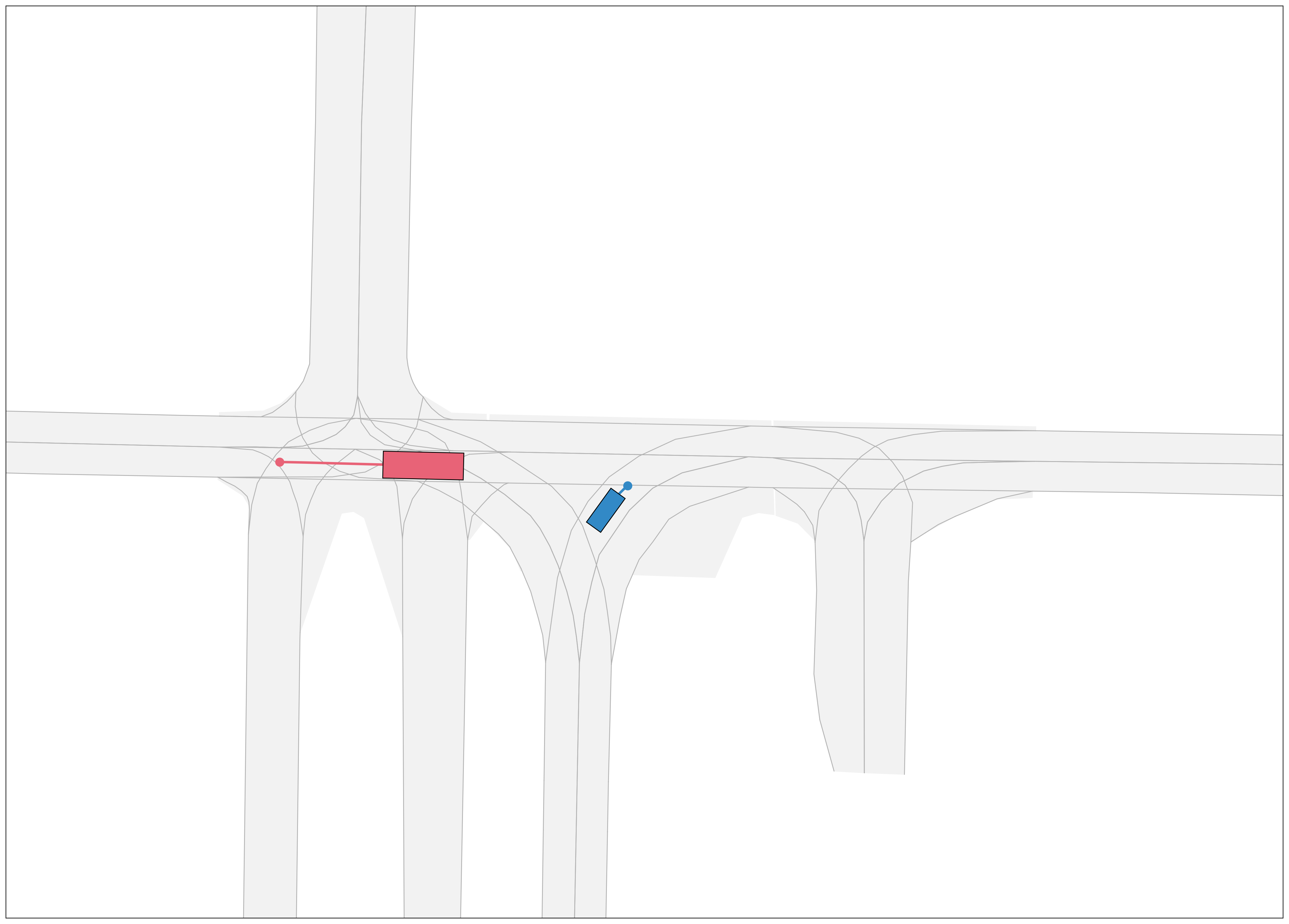}
\captionof{figure}{\textbf{Qualitative Comparison.} We show the baseline
(\textbf{top}) and our approach (\textbf{bottom}) on two different scenario
descriptions. The baseline fails to accelerate past the conflict point. The line behind each actor indicates a 5s history.\protect\footnotemark}
\label{fig:qualitative}
\end{minipage}
\end{figure*}

\SetKwComment{Comment}{/* }{ */}
\SetKw{KwBy}{by}
\begin{algorithm}[t]
    \small
    \caption{Computing the state at a symbolic time}\label{alg:symbolic_time}
    \KwData{$t\in T$, the name of the symbolic time\\
        \qquad\ \ \!$o\in \{0,1,2\}$, the order of the state}
    \KwResult{$f_o(t)$, the expression for the state at time $t$}
    $i \gets T.index(t)$\;
    $terms\_curr \gets []$\;
    \For{$order \leftarrow pieces[i].order$ \KwTo $o$ \KwBy $-1$}{
    $terms \gets []$\;
    \For{$term$ : $terms\_curr$}{
    $terms \gets terms + [antiderivative(term)]$\;
    }
    \eIf{$order == pieces[i].order$ \KwSty{or} $i == 0$}{
    $expr \gets pieces[i].\_rates[order]$\;
    }{
    $expr \gets f_o[i-1]$\;
    }
    $terms \gets terms + [(expr, pieces[i].\_duration, 0)]$\;
    $terms\_curr \gets terms$\;
    }
    $f_o(t) \gets 0$\;
    \For{$term$ : $terms\_curr$}{
        $f_o(t) \gets f_o(t) + term[0]$\;
    }
\end{algorithm}

\SetKwProg{Fn}{Function}{:}{}
\SetKwFunction{FMakeExpr}{MakeExpr}
\begin{algorithm}[t]
    \small
    \caption{Computing the state at a concrete time}\label{alg:concrete_time}
    \KwData{$t \geq 0$, the value of the concrete time\\
        \qquad\ \ \!$o\in \{0,1,2\}$, the order of the state}
    \KwResult{$f_o(t)$, the expression for the state at time $t$}
    \Fn{\FMakeExpr{$pieces$, $rates$, $offset$, $depth$}}{
        \If{$pieces.length() == 0$}{
            \KwRet $rates[o]$\;
        }
        $head \gets pieces[0]$\;
        $tail \gets pieces[1:]$\;
        $lo \gets offset$\;
        $hi \gets offset + head.\_duration$\;
        \eIf{$depth == pieces.length() - 1$}{
            $cond \gets t\ge lo$\;
        }{
            $cond \gets And(t \ge lo, t < hi)$\;
        }
        $then\_expr \gets head_o(t-lo, rates)$\;
        \For{$ord \leftarrow 0$ \KwTo $2$}{
            $rates[ord] \gets head_{ord}(head.\_duration, rates)$\;
        }
        $else\_expr \gets $\FMakeExpr{$tail, rates, hi, depth + 1$}\;
        \KwRet $If(cond$, $then\_expr$, $else\_expr)$\;
    }
    \KwRet\FMakeExpr{$pieces$, $[0,0,0]$, $0$, $0$}\;
\end{algorithm}

\subsection{Scenario Constraints}
Constraints form the key building block of our scenario representation.
We formalize constraints as a set of logical predicates over
our actor trajectory and driving path representation described in \cref{sec:scenario-representation}.
Let
\begin{equation}
    \mathcal{C} = \{c_1, \dots, c_n\}
\end{equation}
be a set of constraints, each of which is a Boolean function $c: \mathcal{T} \times \{\rho_i\}_{i=1}^{N} \mapsto \{0, 1\}$.
A scenario is satisfiable if there exists a configuration such that all constraints in $\mathcal{C}$ hold.

When specifying constraints, standard boolean logic operators, arithmetic, functions, arrays, etc. are supported.
Moreover, \cref{alg:symbolic_time,alg:concrete_time} define
functions for evaluating actor states are both symbolic (unassigned, free variable)
and concrete (assigned, fixed) times respectively. This allows for both constraints like
``\emph{Actor $i$ and actor $j$ must have equal velocity at some time $t$}'' (symbolic time) and
``\emph{Actor $i$ at time $t=3.4$ must have acceleration $a=0.1$}'' (concrete time).
Because we have annoated locations of interest along each actor's path, we can also specify
constraints relating actors to specific locations; e.g.,
``\emph{Actor $i$ must be 10m from the intersection and 5m in front of Actor $j$ at some time $t$}''.
This allows for a flexible way to define scenarios through the composition of
various simple, declarative constraints.
Boilerplate constraints such as kinematic bounds, scenario length, etc.
can be reused and applied to all scenarios. More complex ideas that arise
from multiple constraints can be saved as template and reused as well.

\footnotetext{Note that these results use GPT-5 with a high reasoning effort.}

\subsection{Closed-loop Replanning}
In most cases, the purpose of orchestration is to test the behavior an AV policy that
is not under the control of the orchestrator.
Thus, the orchestrator must be able to react to new AV behavior over the
course of a closed-loop rollout of the scenario.
In our framework, constraints can naturally specify closed-loop behavior, e.g.
``\emph{Actor $i$ maintains the same speed as ego}''.
To support an unknown AV policy, we can simply update the ego's current state
as variable assignments or additional constraints and re-solve at some frequency.
We also found it helpful to add constraints\footnote{e.g. preventing
    the ego from reversing} on the ego's future
motion as well. Intuitively, these constraints act as the solver's model of the
future ego behavior.
Due to our efficient scenario representation, the solve-time itself is relatively quick
compared to the initial constraint generation phase which is either manually
written by humans or completed by an LLM.

\subsection{Foundation model}
While writing constraints manually is still reasonably ergonomic, the ability to specify
scenarios directly in natural language can have many advantages, such as allowing
non-technical team members to directly create scenarios themselves, or potentially
opening up the possibility of leveraging scenario description databases available online.
In this work, we simply prompt an LLM with an in-context example of a language description
of a scenario and the corresponding scenario constraints and program.
We found it helpful to invoke the LLM in separate stages corresponding to different components
of our scenario program. Specifically, we first ask for a pseudocode representation of the constraints
before programming them into Python.
In this work, we use GPT-5\cite{gpt5} to leverage the
Z3~\cite{de2008z3} API to specify constraints before calling the solver,
but our approach is agnostic to the specific LLM or solver.
An example of the scenario description and prompt can be found in the appendix.

%% file: experiments.tex
\section{EXPERIMENTS}

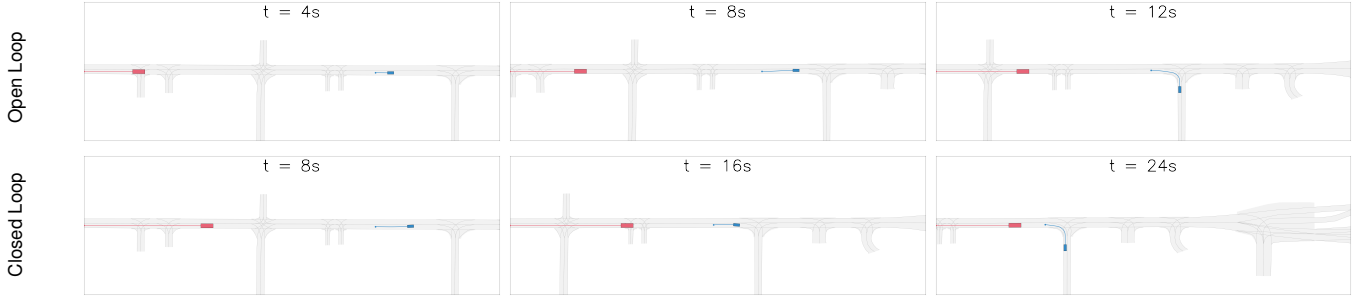
\begin{figure*}[t]
    \centering
    \begin{picture}(25,25)
    \put(0,7){\rotatebox{90}{\textsf{\scriptsize Open Loop}}}
    \end{picture}
    \includegraphics[width=0.31\linewidth]{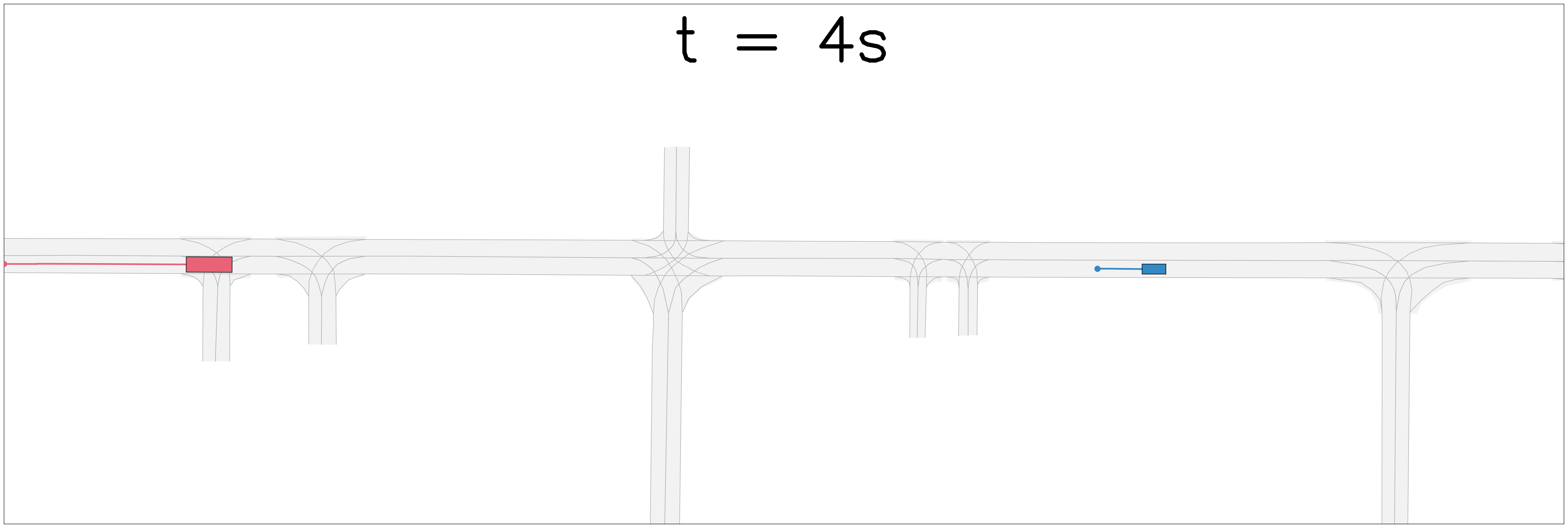}
    \includegraphics[width=0.31\linewidth]{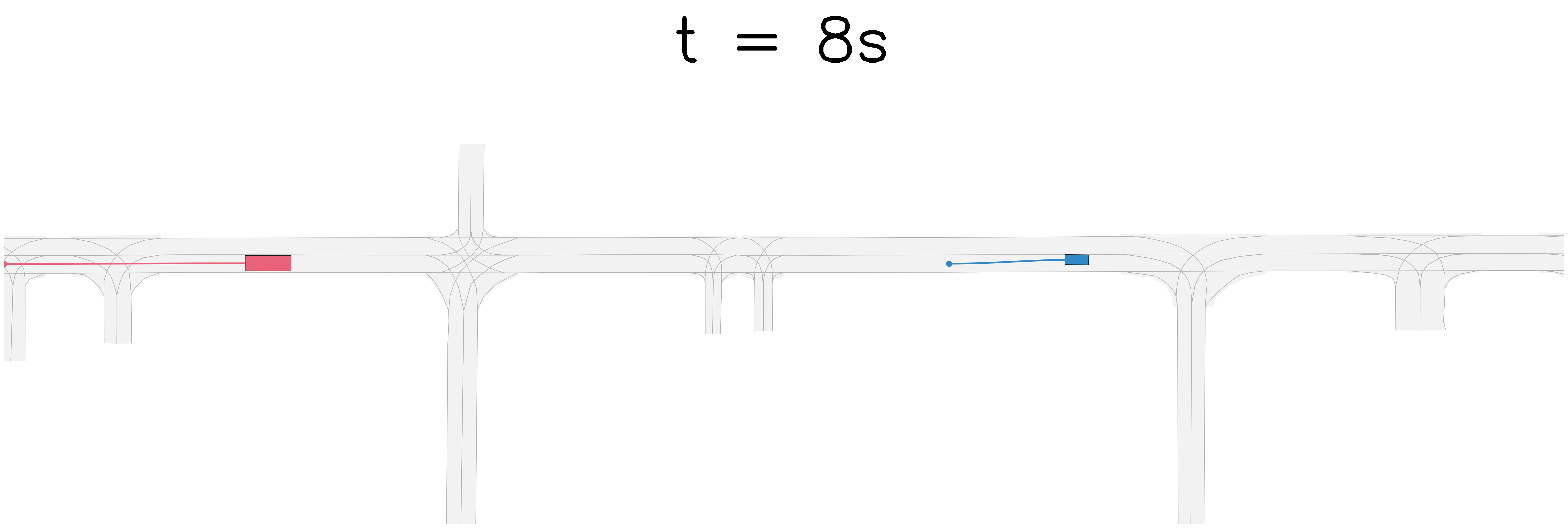}
    \includegraphics[width=0.31\linewidth]{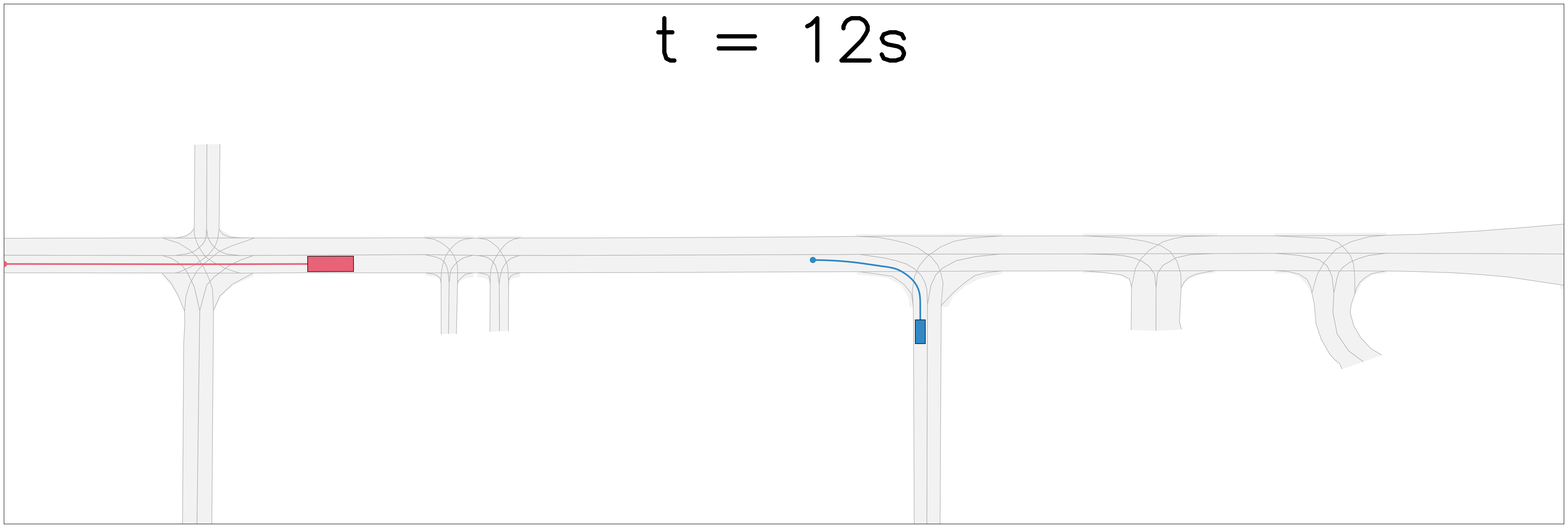}\\\vspace*{.5em}
    \begin{picture}(25,25)
    \put(0,7){\rotatebox{90}{\textsf{\scriptsize Closed Loop}}}
    \end{picture}
    \includegraphics[width=0.31\linewidth]{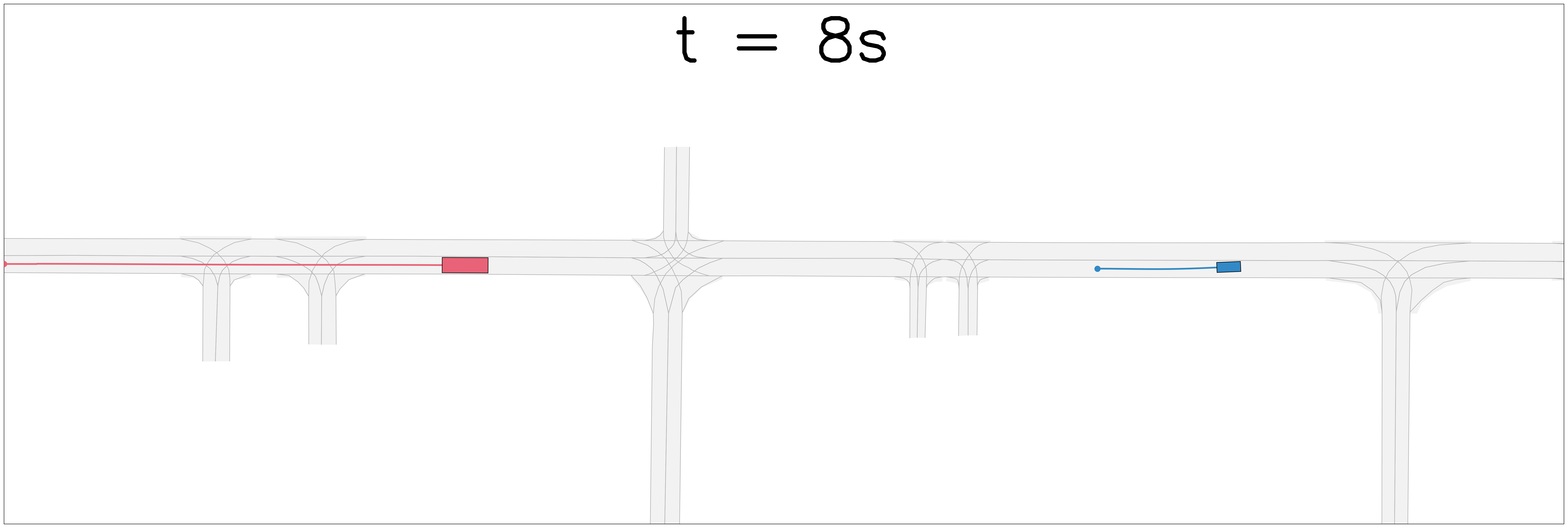}
    \includegraphics[width=0.31\linewidth]{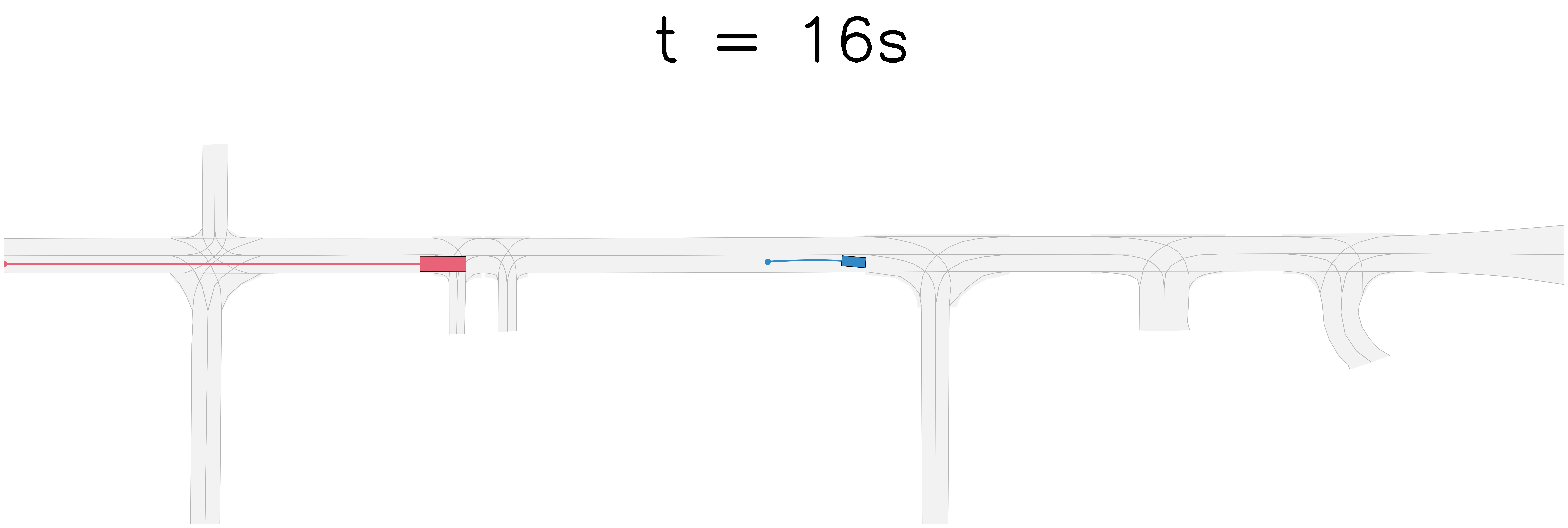}
    \includegraphics[width=0.31\linewidth]{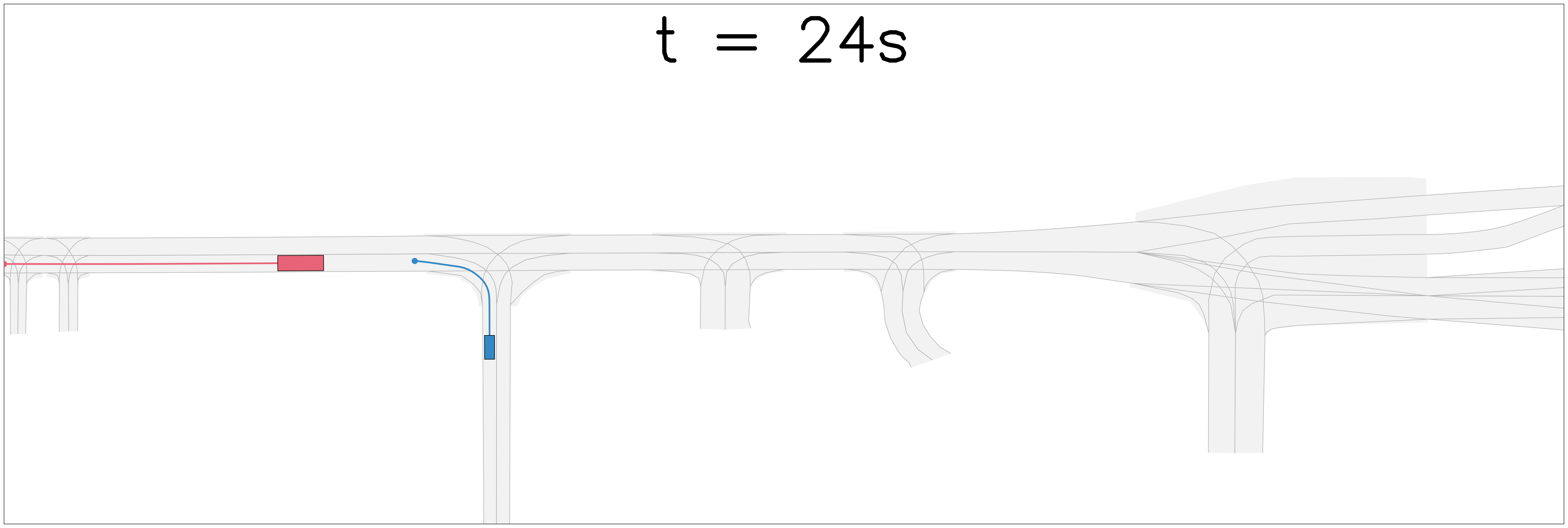}
    \caption{Side-by-side comparison of open loop vs. closed loop orchestration
        for a lead actor turn into driveway scenario. In the open loop mode,
        the lead actor makes the turn while the ego actor remains distant,
        whereas in the closed loop mode, the lead actor makes the turn as
        the ego actor approaches it. (top) Open loop orchestration at every 4s
        with 4s long history. (bottom) Closed loop orchestration at every 8s
        with 8s long history. The lead actor waits for the ego, which
        decelerates due to the lead actor, to approach it in order to meet the
        reactive specification of making the turn 50m ahead of the ego.}
    \label{fig:open-vs-closed-qual}
\end{figure*}
\begin{figure}[t]
    \includegraphics[width=\linewidth]{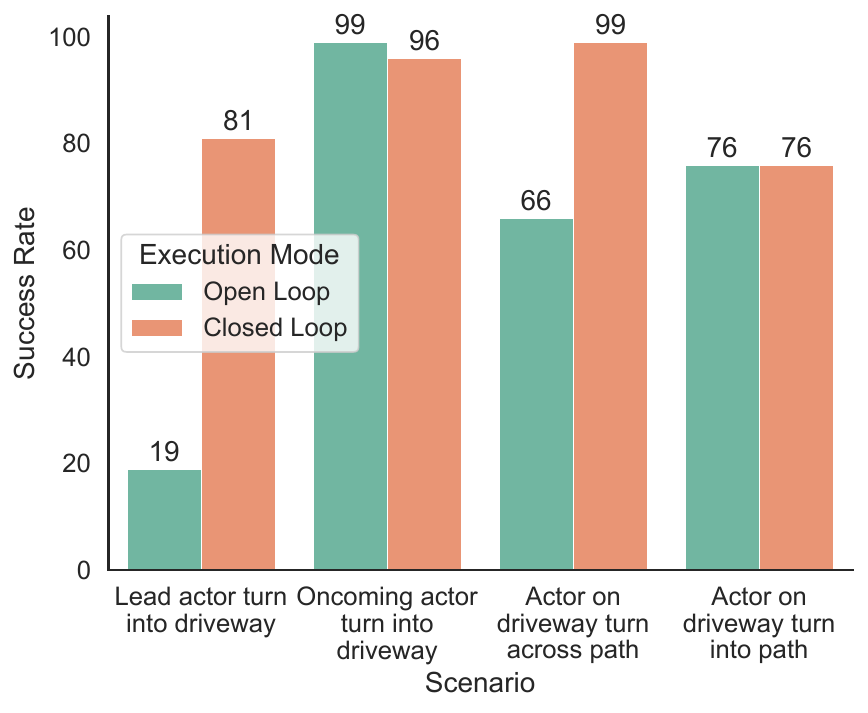}
    \caption{\textbf{Importance of Closed-loop.} We find closed-loop replanning in order to react to
        ego behaviors is important, especially for scenarios that require actor/ego interaction.}
    \label{fig:open-vs-closed-quant}
\end{figure}

\subsection{Experiment Setting}\label{sec:exp-setting}
\subsubsection{Scenario Descriptions}
We conduct experiments on a set of 80 scenario descriptions.
We sample 20 scenarios for development (e.g., prompt engineering) and hold out the rest for testing.
The scenarios describe the interaction between the ego and an actor at an intersection and vary in:
\begin{itemize}
\item \emph{Intended Route:}
    We vary the ego and the actor's starting locations and their intended routes;
    e.g., lane-following along the road, turning left to enter the driveway, turning right to exit the driveway, etc.
    This dictates the conflict geometry between the ego and the actor–whether they are travelling along the parallel lanes,
    converging to the same lane, or crossing each others' lanes.
\item \emph{Desired Interaction:}
    We vary the desired interaction between the ego and the actor;
    e.g., the actor should yield, not yield, or stop-and-go to the ego turning across its path,
    the actor should hard brake to a stop, etc.
\item \emph{Trigger Condition:}
    We vary the trigger condition for the desired interaction;
    e.g., the ego's distance behind the actor, the ego and the actor's distance or time-to-arrival at a conflict point,
    the ego and the actor's distance or time-to-arrival to the intersection, etc.
\end{itemize}

\subsubsection{Metrics}
Our primary metric is orchestration success rate. For each scenario description,
we define the success criteria as a programmatic function of the scenario trace.
Specifically, we implement a function that, given a scenario description and trace,
evaluates whether the scenarios's intended route, desired interaction, and trigger conditions are met.
We use a best-effort approach to ensure that the intent in the descriptions are accurately captured.
Our success criteria is hand-written, but defining this function based on natural language is an interesting and open research question.

Our success criteria are defined with respect to an error tolerance.
For example, if the trigger condition is based on the ego and the actor's distance (resp., time-to-arrival) to a conflict point,
we evaluate whether that trigger condition is met with respect to an error tolerance on the distance (resp., time-to-arrival).
We report success rate over two error tolerance levels.
Overall, this metric allows us to measure the degree of controllability/precision a method is able to offer.

\subsection{Benchmarking Natural Language to Scenario}\label{sec:exp-overall}
We evaluate our method's ability to orchestrate scenarios from natural language descriptions.
In all cases, the ego vehicle is equipped with an AV policy that the orchestrator
does not have control over.
\subsubsection{Baseline}
We use ProSim~\cite{tan2024promptable}, a state-of-the-art deep learning based
conditional traffic simulation model as our baseline,
with some adaptations to our setting.
Firstly, because ProSim does not handle actor initialization and also requires
1 second of state history, we use our method's output at from $t=0$  to $t=1$ to initialize and
warmup ProSim.
Secondly, as ProSim's language conditioning does not handle arbitrary prompts, we found that
the most effective approach was to use our method's plans as intermediate $(x,y,t)$ waypoints
for ProSim conditioning. Specifically, we re-run the solver at 1hz and give ProSim
the waypoint planned $(x, y)$ 4 seconds from the current time.

\subsubsection{Implementation Details}
We use GPT-5 \cite{gpt5} as the LLM. We tried reasoning efforts of low, medium,
and high and found medium to work the best. For the external, off-the-shelf SMT
solver we use Z3 \cite{de2008z3} via the Python API.

\subsubsection{Results}

We show quantitative results in \cref{fig:quantitative} and qualitative results in \cref{fig:qualitative}.
Overall, we find that our approach outperforms the baseline in orchestration success rate.
Despite using the plans from our approach as goal points, the ProSim baseline
has a lower success rate. We found that the most common failure mode was ProSim
not executing the scenario to the precise timing of the solver. For instance, if
a constraint is to meet the ego at a certain conflict point, ProSim may fail to
accelerate fast enough, or in another case, ProSim may fail to go ahead and not
yield. This suggests that while statistical models may be beneficial in
generating more in-distribution scenarios, precise or slightly
out-of-distribution scenarios remain a challenge.

\subsection{Precision \& Controllability}\label{sec:precision}

\begin{figure}[t]
    \includegraphics[width=\linewidth]{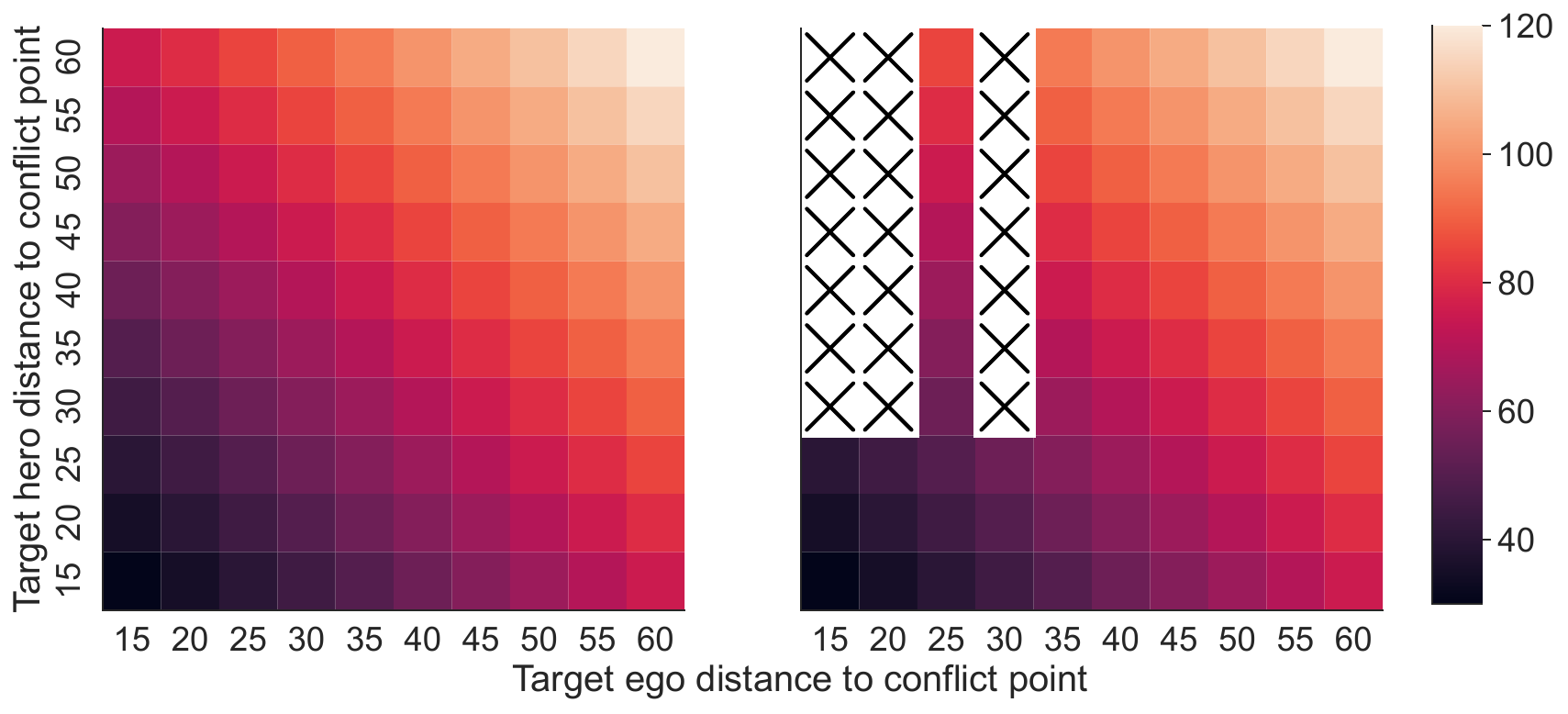}
    \caption{(left) Ours. (right) Baseline. The cell color represents the sum of
    the x and y values. A cell with an ``x'' indicates that the reactive trigger
    was not met. We evaluate both methods at a distance threshold of 2.0m. We
    relax the validation criteria for the baseline so that the actor does not
    have to yield at the specified distance but note that this is part of the
    test intention and that most of the baseline trials fail for this reason.}
    \label{fig:precision}
\end{figure}

To assess the precision and controllability of our method, we analyze its
performance on a fixed scenario across a discretized grid of parameter values in
comparison to the baseline (\cref{fig:precision}).
The scenario description is shown in \hyperref[box:description]{Box 1}.
We find that the baseline systematically fails when the target
ego distance to the conflict point is low. This is problematic for a testing
framework which must have coverage over the range of possible scenario
parameterizations in order to provide a reasonable guarantee on safety. 

\subsection{Open loop vs. Closed loop Execution}\label{sec:exp-loop}

Our next experiment analyzes the importance of our closed-loop approach to scenario orchestration.
Specifically, we evaluate our approach on four representative scenarios with closed-loop replanning
disabled (``open-loop execution'') and enables (``closed-loop execution'').
In both cases, we emphasize that the ego is controlled by a different AV policy executing in closed-loop.
To control for any errors attributable to the LLM, we manually implement constraints for each scenario.
These scenarios span the possible intended routes for the actor at an intersection.
As before, each scenario also contains a trigger condition that relates the ego and non-ego actors
which the scenario framework needs to achieve in closed-loop.
Qualitative and quantitative results are shown in \cref{fig:open-vs-closed-qual,fig:open-vs-closed-quant} respectively.

We find that for half of the scenarios, closed-loop execution makes a
significant difference over open-loop execution.
For the remaining half of the scenarios, we believe
that the open- vs. closed-loop versions are equivalent due to the semantics of
the scenarios involving a trigger point that does not change over time and thus
the scenario is not sensitive to the behavior of the ego vehicle. Due to the
complexity of closed-loop, some errors may result such as a failure to solve the
constraint satisfaction problem, leading to the slightly worse performance of
closed-loop on the oncoming actor scenario. Overall, closed-loop orchestration
achieves a 23\% higher success rate over open-loop orchestration.

\subsection{Runtime Analysis}
\begin{figure}[t]
    \includegraphics[width=\linewidth]{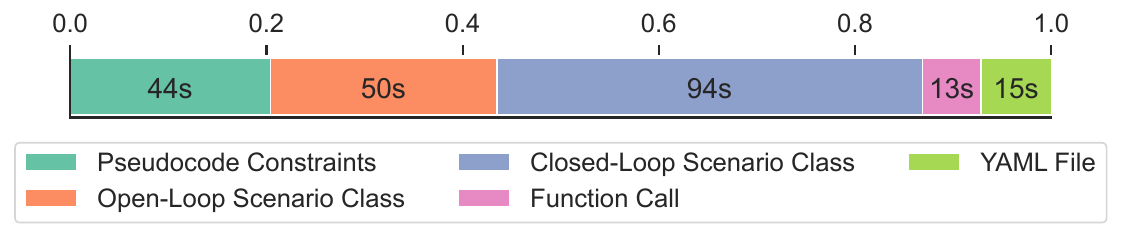}
    \caption{\textbf{LLM Generation Runtime.} We measure the average runtime of
    the one-time LLM generation phase. In our experiments, we use GPT-5 (medium
   reasoning effort).}
    \label{fig:generation-runtime}
\end{figure}
\begin{figure}[t]
    \includegraphics[width=\linewidth]{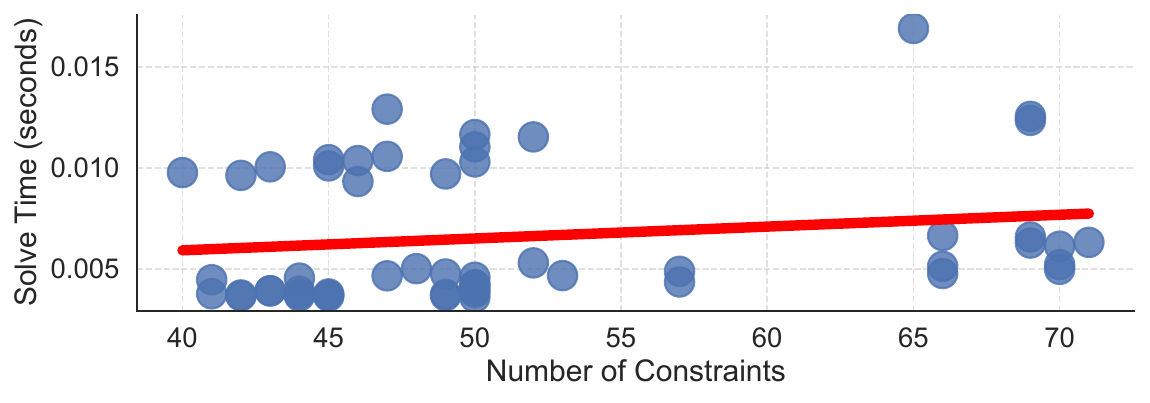}
    \caption{\textbf{Solver Runtime.} We measure the runtime of the solver when given varying number of constraints.}
    \label{fig:solver-runtime}
\end{figure}

In \cref{fig:generation-runtime,fig:solver-runtime}, we analyze the runtime of our approach.
Overall, the one-time LLM generation phase takes on the order of minutes while
the solver, which is run in closed loop at the re-planning frequency, runs
within a fraction of a second. During closed loop execution, we solve at
various tolerances until a solution is found, or raise a failure at the maximum
tolerance, with a solve timeout of 10s. Solving can take longer than a few
seconds during closed loop execution depending on the difficulty of the problem.

%% file: conclusion.tex
\section{CONCLUSIONS \& DISCUSSION}
In this paper we introduced a neurosymbolic framework for 
closed-loop traffic scenario orchestration
By delegating high-level spatial-temporal reasoning 
and natural-language understanding to an LLM, 
and precise continuous-time motion planning to an SMT solver, 
our method achieves controllable, reactive, and physically 
realizable scenarios from free-form textual specifications.
Our experiments on a diverse set of scenario descriptions 
show that this hybrid design consistently outperforms a 
state-of-the-art learning-only baseline, 
particularly for multi-actor interactions and edge-case conditions 
that have precise spatial temporal requirements.
The solver's ability to re-plan in closed loop enables 
seamless adaptation to the ego vehicle's evolving trajectory, 
especially important in scenarios with reactive constraints.

We now discuss some limitations.
Our use of an LLM means that errors such as hallucinations can propagate through
our system. Our implementation features a single hero actor and map which should
be scaled up to greater real-world complexity. 
Lastly, our experiments did not measure statistical realism.
Measuring realism is difficult since our descriptions do not correspond to any existing
ground truth execution. Most realism metrics are based on data likelihood, but
we are also not focused on orchestrating the most likely scenarios. Thus, as a proxy, we
have resorted to ensuring all methods create kinematically feasible trajectories.
However, exploring additional metrics and methods which consider statistical realism is a promising line of future work.

%% file: appendix.tex
\section*{APPENDIX}

\subsection{Supplementary Algorithms}

We include subroutines referenced in \cref{alg:symbolic_time,alg:concrete_time}.

\SetKwComment{Comment}{/* }{ */}
\begin{algorithm}[h]
    \small
    \caption{Calculating the antiderivative of a term}\label{alg:antiderivative}
    \KwData{$expr$, the term expression\\
        \qquad\ \ \!$var$, the variable that the term is with respect to\\
        \qquad\ \ \!$o\in \{0,1,2\}$, the order of the term}
    \KwResult{$\int expr$, the antiderivative of the term\\
    \qquad\ \ \ \ \!$o$, the next order of the term
    }
    $\int expr \gets \tfrac{max(1, o)}{o + 1} \cdot expr \cdot var$\;
    $o \gets o + 1$\;
\end{algorithm}

\begin{algorithm}[h]
    \small
    \caption{Computing the piece at a time}\label{alg:piece}
    \KwData{$t$, the time\\
        \qquad\ \ \!$o\in \{0,1,2\}$, the order of the state\\
        \qquad\ \ \!$piece$, the piece\\
        \qquad\ \ \!$rates$, the cumulative value at each order so far\\
        }
    \KwResult{$piece_o(t, rates)$, the expression for the piece at time $t$}
    $s \gets 0$\;
    \eIf{$rates$}{
      $rates\_next \gets []$\;
      \For{$order \leftarrow 0$ \KwTo $piece.order + 1$}{
        $\_r \gets piece.\_rates[order]$\;
        $r \gets rates[order]$\;
        \eIf{$order < piece.order$}{
          $rates\_next \gets rates\_next + [\_r + r]$\;
        }{
          $rates\_next \gets rates\_next + [\_r]$\;
        }
      }
      $rates \gets rates\_next$\;
    }{
      $rates \gets piece.\_rates$\;
    }
    \For{$k \leftarrow o$ \KwTo $piece.order + 1$}{
      $power \gets k - o$\;
      \eIf{power}{
        $val \gets t^{power}$\;
      }{
        $val \gets 1$\;
      }
      $s \gets s + \tfrac{1}{power!}\cdot rates[k]\cdot val$\;
    }
\end{algorithm}

\subsection{Prompts}

We include excerpts of our prompts for multiple stages of our LLM-based generation.
\vspace{5pt}

\noindent
\begin{minipage}{\linewidth}
  \begin{tcolorbox}[title={\textbf{\scriptsize Natural Language to Constraint Generation Prompt}}, enhanced, breakable=false, colback=white, colframe=black, fontupper=\scriptsize]
    You convert descriptions of surface street driving scenarios into actor trajectories and pseudocode constraints.
    The scenario consists of exactly two actors, one ego vehicle (id = 0) and one other actor (id = 1).

    \shortened

    \# Pseudocode constraints

    To refer to the acceleration of an actor with id `i` at knot `t\{j\}`, use `A\{i\}a("t\{j\}")`.
    To refer
    \shortened

  \end{tcolorbox}
  \vspace{5pt}
\end{minipage}
\noindent
\begin{minipage}{\linewidth}
  \begin{tcolorbox}[title={\textbf{\scriptsize Constraint to Scenario Class Generation Prompt}}, enhanced, breakable=false, colback=white, colframe=black, fontupper=\scriptsize]
    You translate actor trajectories and pseudocode constraints into a Python class.

    The Python class should inherit from the Scenario class and be named the scenario name converted to camel case and with "Scenario" appended.

    \shortened

    In the pseudocode constraints, there are special constants `stop\_line`, `conflict\_point`, `turn\_start`, and `turn\_end`.
    \shortened

  \end{tcolorbox}
  \vspace{5pt}
\end{minipage}
\noindent
\begin{minipage}{\linewidth}
  \begin{tcolorbox}[title={\textbf{\scriptsize Open Loop to Closed Loop Scenario Class Generation Prompt}}, enhanced, breakable=false, colback=white, colframe=black, fontupper=\scriptsize]
    Modify the given Scenario class as follows:

    - add self.search = True in the .reset() method

    \shortened

    Output the complete, modified Scenario class after making all the requested modifications. Do not output anything else.
  \end{tcolorbox}
\end{minipage}

%% file: appendix_extended.tex
\newpage
\onecolumn
\section{Scenario Taxonomy}
\label{sec:taxonomy}

\setlength{\LTleft}{\fill}
\setlength{\LTright}{\fill}
\begin{longtable}{llllll}
\textbf{Ego State} & \textbf{Ego Intent} & \textbf{Hero State} & \textbf{Hero Intent} & \textbf{Interaction}  & \textbf{Trigger} \\
\endfirsthead
\textbf{Ego State} & \textbf{Ego Intent} & \textbf{Hero State} & \textbf{Hero Intent} & \textbf{Interaction}  & \textbf{Trigger} \\
\endhead
Driveway           & Right turn          & Driveway            & Right turn           & Hero brakes           & Distance         \\
Driveway           & Right turn          & Westbound           & Lane follow          & Hero yields           & Distance         \\
Driveway           & Right turn          & Westbound           & Lane follow          & Hero does not yield   & Distance         \\
Driveway           & Right turn          & Westbound           & Lane follow          & Hero yields then goes & Distance         \\
Driveway           & Left turn           & Driveway            & Left turn            & Hero brakes           & Distance         \\
Driveway           & Left turn           & Westbound           & Lane follow          & Hero yields           & Distance         \\
Driveway           & Left turn           & Westbound           & Lane follow          & Hero does not yield   & Distance         \\
Driveway           & Left turn           & Westbound           & Lane follow          & Hero yields then goes & Distance         \\
Driveway           & Left turn           & Eastbound           & Lane follow          & Hero yields           & Distance         \\
Driveway           & Left turn           & Eastbound           & Lane follow          & Hero does not yield   & Distance         \\
Driveway           & Left turn           & Eastbound           & Lane follow          & Hero yields then goes & Distance         \\
Driveway           & Left turn           & Eastbound           & Left turn            & Hero yields           & Distance         \\
Driveway           & Left turn           & Eastbound           & Left turn            & Hero does not yield   & Distance         \\
Driveway           & Left turn           & Eastbound           & Left turn            & Hero yields then goes & Distance         \\
Westbound          & Lane follow         & Westbound           & Right turn           & Hero brakes           & Distance         \\
Westbound          & Lane follow         & Eastbound           & Left turn            & Hero yields           & Distance         \\
Westbound          & Lane follow         & Eastbound           & Left turn            & Hero does not yield   & Distance         \\
Westbound          & Lane follow         & Eastbound           & Left turn            & Hero yields then goes & Distance         \\
Westbound          & Lane follow         & Driveway            & Right turn           & Hero yields           & Distance         \\
Westbound          & Lane follow         & Driveway            & Right turn           & Hero does not yield   & Distance         \\
Westbound          & Lane follow         & Driveway            & Right turn           & Hero yields then goes & Distance         \\
Westbound          & Lane follow         & Driveway            & Left turn            & Hero yields           & Distance         \\
Westbound          & Lane follow         & Driveway            & Left turn            & Hero does not yield   & Distance         \\
Westbound          & Lane follow         & Driveway            & Left turn            & Hero yields then goes & Distance         \\
Westbound          & Right turn          & Eastbound           & Left turn            & Hero yields           & Distance         \\
Westbound          & Right turn          & Eastbound           & Left turn            & Hero does not yield   & Distance         \\
Westbound          & Right turn          & Eastbound           & Left turn            & Hero yields then goes & Distance         \\
Eastbound          & Lane follow         & Eastbound           & Left turn            & Hero brakes           & Distance         \\
Eastbound          & Lane follow         & Driveway            & Left turn            & Hero yields           & Distance         \\
Eastbound          & Lane follow         & Driveway            & Left turn            & Hero does not yield   & Distance         \\
Eastbound          & Lane follow         & Driveway            & Left turn            & Hero yields then goes & Distance         \\
Eastbound          & Left turn           & Westbound           & Lane follow          & Hero yields           & Distance         \\
Eastbound          & Left turn           & Westbound           & Lane follow          & Hero does not yield   & Distance         \\
Eastbound          & Left turn           & Westbound           & Lane follow          & Hero yields then goes & Distance         \\
Eastbound          & Left turn           & Westbound           & Right turn           & Hero yields           & Distance         \\
Eastbound          & Left turn           & Westbound           & Right turn           & Hero does not yield   & Distance         \\
Eastbound          & Left turn           & Westbound           & Right turn           & Hero yields then goes & Distance         \\
Eastbound          & Left turn           & Driveway            & Left turn            & Hero yields           & Distance         \\
Eastbound          & Left turn           & Driveway            & Left turn            & Hero does not yield   & Distance         \\
Eastbound          & Left turn           & Driveway            & Left turn            & Hero yields then goes & Distance         \\
Driveway           & Right turn          & Driveway            & Right turn           & Hero brakes           & Time             \\
Driveway           & Right turn          & Westbound           & Lane follow          & Hero yields           & Time             \\
Driveway           & Right turn          & Westbound           & Lane follow          & Hero does not yield   & Time             \\
Driveway           & Right turn          & Westbound           & Lane follow          & Hero yields then goes & Time             \\
Driveway           & Left turn           & Driveway            & Left turn            & Hero brakes           & Time             \\
Driveway           & Left turn           & Westbound           & Lane follow          & Hero yields           & Time             \\
Driveway           & Left turn           & Westbound           & Lane follow          & Hero does not yield   & Time             \\
Driveway           & Left turn           & Westbound           & Lane follow          & Hero yields then goes & Time             \\
Driveway           & Left turn           & Eastbound           & Lane follow          & Hero yields           & Time             \\
Driveway           & Left turn           & Eastbound           & Lane follow          & Hero does not yield   & Time             \\
Driveway           & Left turn           & Eastbound           & Lane follow          & Hero yields then goes & Time             \\
Driveway           & Left turn           & Eastbound           & Left turn            & Hero yields           & Time             \\
Driveway           & Left turn           & Eastbound           & Left turn            & Hero does not yield   & Time             \\
Driveway           & Left turn           & Eastbound           & Left turn            & Hero yields then goes & Time             \\
Westbound          & Lane follow         & Westbound           & Right turn           & Hero brakes           & Time             \\
Westbound          & Lane follow         & Eastbound           & Left turn            & Hero yields           & Time             \\
Westbound          & Lane follow         & Eastbound           & Left turn            & Hero does not yield   & Time             \\
Westbound          & Lane follow         & Eastbound           & Left turn            & Hero yields then goes & Time             \\
Westbound          & Lane follow         & Driveway            & Right turn           & Hero yields           & Time             \\
Westbound          & Lane follow         & Driveway            & Right turn           & Hero does not yield   & Time             \\
Westbound          & Lane follow         & Driveway            & Right turn           & Hero yields then goes & Time             \\
Westbound          & Lane follow         & Driveway            & Left turn            & Hero yields           & Time             \\
Westbound          & Lane follow         & Driveway            & Left turn            & Hero does not yield   & Time             \\
Westbound          & Lane follow         & Driveway            & Left turn            & Hero yields then goes & Time             \\
Westbound          & Right turn          & Eastbound           & Left turn            & Hero yields           & Time             \\
Westbound          & Right turn          & Eastbound           & Left turn            & Hero does not yield   & Time             \\
Westbound          & Right turn          & Eastbound           & Left turn            & Hero yields then goes & Time             \\
Eastbound          & Lane follow         & Eastbound           & Left turn            & Hero brakes           & Time             \\
Eastbound          & Lane follow         & Driveway            & Left turn            & Hero yields           & Time             \\
Eastbound          & Lane follow         & Driveway            & Left turn            & Hero does not yield   & Time             \\
Eastbound          & Lane follow         & Driveway            & Left turn            & Hero yields then goes & Time             \\
Eastbound          & Left turn           & Westbound           & Lane follow          & Hero yields           & Time             \\
Eastbound          & Left turn           & Westbound           & Lane follow          & Hero does not yield   & Time             \\
Eastbound          & Left turn           & Westbound           & Lane follow          & Hero yields then goes & Time             \\
Eastbound          & Left turn           & Westbound           & Right turn           & Hero yields           & Time             \\
Eastbound          & Left turn           & Westbound           & Right turn           & Hero does not yield   & Time             \\
Eastbound          & Left turn           & Westbound           & Right turn           & Hero yields then goes & Time             \\
Eastbound          & Left turn           & Driveway            & Left turn            & Hero yields           & Time             \\
Eastbound          & Left turn           & Driveway            & Left turn            & Hero does not yield   & Time             \\
Eastbound          & Left turn           & Driveway            & Left turn            & Hero yields then goes & Time            
\end{longtable}

\section{Scenario Descriptions}
\label{sec:benchmark}

\begin{enumerate}
    \item Ego is on the driveway exit and it is initial\_ego\_distance\_behind\_stop\_line\_m behind the stop line. It wants to turn right onto the road travelling westbound. There is a hero vehicle on the driveway exit and it is initial\_hero\_distance\_behind\_stop\_line\_m behind the stop line. It wants to turn right onto the road travelling westbound. When the ego is ego\_distance\_behind\_hero\_m behind the hero vehicle, the hero vehicle begins to decelerate to a complete stop at a rate of hero\_deceleration\_mpss.
    \item Ego is on the driveway exit and it is initial\_ego\_distance\_behind\_stop\_line\_m behind the stop line. It wants to turn right onto the road travelling westbound. There is a hero vehicle on the road. It wants to lane follow along the road travelling westbound. When the ego is ego\_distance\_behind\_stop\_line\_m behind the stop line and the hero vehicle is hero\_distance\_behind\_conflict\_point\_m behind the conflict point, the hero vehicle begins to decelerate to a complete stop at end\_hero\_distance\_behind\_conflict\_point\_m behind the conflict point.
    \item Ego is on the driveway exit and it is initial\_ego\_distance\_behind\_stop\_line\_m behind the stop line. It wants to turn right onto the road travelling westbound. There is a hero vehicle on the road. It wants to lane follow along the road travelling westbound. When the ego is ego\_distance\_behind\_stop\_line\_m behind the stop line and the hero vehicle is distance\_behind\_conflict\_point\_m behind the conflict point, the hero vehicle begins to accelerate to pass the conflict point before the ego begins to turn.
    \item Ego is on the driveway exit and it is initial\_ego\_distance\_behind\_stop\_line\_m behind the stop line. It wants to turn right onto the road travelling westbound. There is a hero vehicle on the road. It wants to lane follow along the road travelling westbound. When the ego is ego\_distance\_behind\_stop\_line\_m behind the stop line and the hero vehicle is distance\_behind\_conflict\_point\_m behind the conflict point, the hero vehicle begins to decelerate to a complete stop at end\_hero\_distance\_behind\_conflict\_point\_m behind the conflict point. Then, the hero vehicle begins to accelerate to pass the conflict point before the ego begins to turn.
    \item Ego is on the driveway exit and it is initial\_ego\_distance\_behind\_stop\_line\_m behind the stop line. It wants to turn left onto the road travelling eastbound. There is a hero vehicle on the driveway exit and it is initial\_hero\_distance\_behind\_stop\_line\_m behind the stop line. It wants to turn left onto the road travelling eastbound. When the ego is ego\_distance\_behind\_hero\_m behind the hero vehicle, the hero vehicle begins to decelerate to a complete stop at a rate of hero\_deceleration\_mpss.
    \item Ego is on the driveway exit and it is initial\_ego\_distance\_behind\_stop\_line\_m behind the stop line. It wants to turn left onto the road travelling eastbound. There is a hero vehicle on the road. It wants to lane follow along the road travelling westbound. When the ego is ego\_distance\_behind\_stop\_line\_m behind the stop line and the hero vehicle is hero\_distance\_behind\_conflict\_point\_m behind the conflict point, the hero vehicle begins to decelerate to a complete stop at end\_hero\_distance\_behind\_conflict\_point\_m behind the conflict point.
    \item Ego is on the driveway exit and it is initial\_ego\_distance\_behind\_stop\_line\_m behind the stop line. It wants to turn left onto the road travelling eastbound. There is a hero vehicle on the road. It wants to lane follow along the road travelling westbound.  When the ego is ego\_distance\_behind\_stop\_line\_m behind the stop line and the hero vehicle is distance\_behind\_conflict\_point\_m behind the conflict point, the hero vehicle begins to accelerate to pass the conflict point before the ego begins to turn.
    \item Ego is on the driveway exit and it is initial\_ego\_distance\_behind\_stop\_line\_m behind the stop line. It wants to turn left onto the road travelling eastbound. There is a hero vehicle on the road. It wants to lane follow along the road travelling westbound.  When the ego is ego\_distance\_behind\_stop\_line\_m behind the stop line and the hero vehicle is distance\_behind\_conflict\_point\_m behind the conflict point, the hero vehicle begins to decelerate to a complete stop at end\_hero\_distance\_behind\_conflict\_point\_m behind the conflict point. Then, the hero vehicle begins to accelerate to pass the conflict point before the ego begins to turn.
    \item Ego is on the driveway exit and it is initial\_ego\_distance\_behind\_stop\_line\_m behind the stop line. It wants to turn left onto the road travelling eastbound. There is a hero vehicle on the road. It wants to lane follow along the road travelling eastbound. When the ego is ego\_distance\_behind\_stop\_line\_m behind the stop line and the hero vehicle is hero\_distance\_behind\_conflict\_point\_m behind the conflict point, the hero vehicle begins to decelerate to a complete stop at end\_hero\_distance\_behind\_conflict\_point\_m behind the conflict point.
    \item Ego is on the driveway exit and it is initial\_ego\_distance\_behind\_stop\_line\_m behind the stop line. It wants to turn left onto the road travelling eastbound. There is a hero vehicle on the road. It wants to lane follow along the road travelling eastbound. When the ego is ego\_distance\_behind\_stop\_line\_m behind the stop line and the hero vehicle is distance\_behind\_conflict\_point\_m behind the conflict point, the hero vehicle begins to accelerate to pass the conflict point before the ego begins to turn.
    \item Ego is on the driveway exit and it is initial\_ego\_distance\_behind\_stop\_line\_m behind the stop line. It wants to turn left onto the road travelling eastbound. There is a hero vehicle on the road. It wants to lane follow along the road travelling eastbound. When the ego is ego\_distance\_behind\_stop\_line\_m behind the stop line and the hero vehicle is distance\_behind\_conflict\_point\_m behind the conflict point, the hero vehicle begins to decelerate to a complete stop at end\_hero\_distance\_behind\_conflict\_point\_m behind the conflict point. Then, the hero vehicle begins to accelerate to pass the conflict point before the ego begins to turn.
    \item Ego is on the driveway exit and it is initial\_ego\_distance\_behind\_stop\_line\_m behind the stop line. It wants to turn left onto the road travelling eastbound. There is a hero vehicle on the road. It wants to turn left into the driveway. When the ego is ego\_distance\_behind\_stop\_line\_m behind the stop line and the hero vehicle is hero\_distance\_behind\_conflict\_point\_m behind the conflict point, the hero vehicle begins to decelerate to a complete stop at end\_hero\_distance\_behind\_conflict\_point\_m behind the conflict point.
    \item Ego is on the driveway exit and it is initial\_ego\_distance\_behind\_stop\_line\_m behind the stop line. It wants to turn left onto the road travelling eastbound. There is a hero vehicle on the road. It wants to turn left into the driveway. When the ego is ego\_distance\_behind\_stop\_line\_m behind the stop line and the hero vehicle is distance\_behind\_conflict\_point\_m behind the conflict point, the hero vehicle begins to accelerate to pass the conflict point before the ego begins to turn.
    \item Ego is on the driveway exit and it is initial\_ego\_distance\_behind\_stop\_line\_m behind the stop line. It wants to turn left onto the road travelling eastbound. There is a hero vehicle on the road. It wants to turn left into the driveway. When the ego is ego\_distance\_behind\_stop\_line\_m behind the stop line and the hero vehicle is distance\_behind\_conflict\_point\_m behind the conflict point, the hero vehicle begins to decelerate to a complete stop at end\_hero\_distance\_behind\_conflict\_point\_m behind the conflict point. Then, the hero vehicle begins to accelerate to pass the conflict point before the ego begins to turn.
    \item Ego is on the road and it is initial\_ego\_distance\_behind\_intersection\_m behind the intersection. It wants to lane follow along the road travelling westbound. There is a hero vehicle on the road. It wants to lane follow along the road travelling westbound. When the ego is ego\_distance\_behind\_hero\_m behind the hero vehicle, the hero vehicle begins to decelerate to a complete stop at a rate of hero\_deceleration\_mpss.
    \item Ego is on the road and it is initial\_ego\_distance\_behind\_intersection\_m behind the intersection. It wants to lane follow along the road travelling westbound. There is a hero vehicle on the road. It wants to turn left into the driveway. When the ego is ego\_distance\_behind\_conflict\_point\_m behind the the conflict point and the hero vehicle is hero\_distance\_behind\_conflict\_point\_m behind the conflict point, the hero vehicle begins to decelerate to a complete stop at end\_hero\_distance\_behind\_conflict\_point\_m behind the conflict point.
    \item Ego is on the road and it is initial\_ego\_distance\_behind\_intersection\_m behind the intersection. It wants to lane follow along the road travelling westbound. There is a hero vehicle on the road. It wants to turn left into the driveway. When the ego is ego\_distance\_behind\_conflict\_point\_m behind the conflict point and the hero vehicle is hero\_distance\_behind\_conflict\_point\_m behind the conflict point, the hero vehicle begins to accelerate to pass the conflict point before the ego begins to turn.
    \item Ego is on the road and it is initial\_ego\_distance\_behind\_intersection\_m behind the intersection. It wants to lane follow along the road travelling westbound. There is a hero vehicle on the road. It wants to turn left into the driveway. When the ego is ego\_distance\_behind\_conflict\_point\_m behind the conflict point and the hero vehicle is hero\_distance\_behind\_conflict\_point\_m behind the conflict point, the hero vehicle begins to decelerate to a complete stop at end\_hero\_distance\_behind\_conflict\_point\_m behind the conflict point. Then, the hero vehicle begins to accelerate to pass the conflict point before the ego passes the conflict point.
    \item Ego is on the road and it is initial\_ego\_distance\_behind\_intersection\_m behind the intersection. It wants to lane follow along the road travelling westbound. There is a hero vehicle on the driveway. It wants to turn right onto the road travelling westbound. When the ego is ego\_distance\_behind\_conflict\_point\_m behind the conflict point and the hero vehicle is hero\_distance\_behind\_stop\_line\_m behind the stop line, the hero vehicle begins to decelerate to a complete stop just before the stop line. 
    \item Ego is on the road and it is initial\_ego\_distance\_behind\_intersection\_m behind the intersection. It wants to lane follow along the road travelling westbound. There is a hero vehicle on the driveway. It wants to turn right onto the road travelling westbound. When the ego is ego\_distance\_behind\_conflict\_point\_m behind the conflict point and the hero vehicle is hero\_distance\_behind\_stop\_line\_m behind the stop line, the hero vehicle begins to accelerate to complete the turn before the ego passes the conflict point.
    \item Ego is on the road and it is initial\_ego\_distance\_behind\_intersection\_m behind the intersection. It wants to lane follow along the road travelling westbound. There is a hero vehicle on the driveway. It wants to turn right onto the road travelling westbound. When the ego is ego\_distance\_behind\_conflict\_point\_m behind the conflict point and the hero vehicle is hero\_distance\_behind\_stop\_line\_m behind the stop line, the hero vehicle begins to decelerate to a complete stop just before the stop line.  Then, the hero vehicle begins to accelerate to complete the turn before the ego passes the conflict point.
    \item Ego is on the road and it is initial\_ego\_distance\_behind\_intersection\_m behind the intersection. It wants to lane follow along the road travelling westbound. There is a hero vehicle on the driveway. It wants to turn left onto the road travelling eastbound. When the ego is ego\_distance\_behind\_conflict\_point\_m behind the conflict point and the hero vehicle is hero\_distance\_behind\_stop\_line\_m behind the stop line, the hero vehicle begins to decelerate to a complete stop just before the stop line. 
    \item Ego is on the road and it is initial\_ego\_distance\_behind\_intersection\_m behind the intersection. It wants to lane follow along the road travelling westbound. There is a hero vehicle on the driveway. It wants to turn left onto the road travelling eastbound. When the ego is ego\_distance\_behind\_conflict\_point\_m behind the conflict point and the hero vehicle is hero\_distance\_behind\_stop\_line\_m behind the stop line, the hero vehicle begins to accelerate to complete the turn before the ego passes the conflict point.
    \item Ego is on the road and it is initial\_ego\_distance\_behind\_intersection\_m behind the intersection. It wants to lane follow along the road travelling westbound. There is a hero vehicle on the driveway. It wants to turn left onto the road travelling eastbound. When the ego is ego\_distance\_behind\_conflict\_point\_m behind the conflict point and the hero vehicle is hero\_distance\_behind\_stop\_line\_m behind the stop line, the hero vehicle begins to decelerate to a complete stop just before the stop line.  Then, the hero vehicle begins to accelerate to complete the turn before the ego passes the conflict point.
    \item Ego is on the road and it is initial\_ego\_distance\_behind\_intersection\_m behind the intersection. It wants to turn right into the driveway. There is a hero vehicle on the road. It wants to turn left into the driveway. When the ego is ego\_distance\_behind\_conflict\_point\_m behind the conflict point and the hero vehicle is hero\_distance\_behind\_conflict\_point\_m behind the conflict point, the hero vehicle begins to decelerate to a complete stop at end\_hero\_distance\_behind\_conflict\_point\_m behind the conflict point.
    \item Ego is on the road and it is initial\_ego\_distance\_behind\_intersection\_m behind the intersection. It wants to lane follow along the road travelling westbound. There is a hero vehicle on the road. It wants to turn left into the driveway. When the ego is ego\_distance\_behind\_conflict\_point\_m behind the conflict point and the hero vehicle is hero\_distance\_behind\_conflict\_point\_m behind the conflict point, the hero vehicle begins to accelerate to pass the conflict point before the ego passes the conflict point.
    \item Ego is on the road and it is initial\_ego\_distance\_behind\_intersection\_m behind the intersection. It wants to lane follow along the road travelling westbound. There is a hero vehicle on the road. It wants to turn left into the driveway. When the ego is ego\_distance\_behind\_conflict\_point\_m behind the conflict point and the hero vehicle is distance\_behind\_conflict\_point\_m behind the conflict point, the hero vehicle begins to decelerate to a complete stop at end\_hero\_distance\_behind\_conflict\_point\_m behind the conflict point. Then, the hero vehicle begins to accelerate to pass the conflict point before the ego passes the conflict point.
    \item Ego is on the road and it is initial\_ego\_distance\_behind\_intersection\_m behind the intersection. It wants to lane follow along the road travelling eastbound. There is a hero vehicle on the road. It wants to lane follow along the road travelling eastbound. When the ego is ego\_distance\_behind\_hero\_m behind the hero vehicle, the hero vehicle begins to decelerate to a complete stop at a rate of hero\_deceleration\_mpss.
    \item Ego is on the road and it is initial\_ego\_distance\_behind\_intersection\_m behind the intersection. It wants to lane follow along the road travelling eastbound. There is a hero vehicle on the driveway. It wants to turn left onto the road travelling eastbound. When the ego is ego\_distance\_behind\_conflict\_point\_m behind the conflict point and the hero vehicle is hero\_distance\_behind\_stop\_line\_m behind the stop line, the hero vehicle begins to decelerate to a complete stop just before the stop line. 
    \item Ego is on the road and it is initial\_ego\_distance\_behind\_intersection\_m behind the intersection. It wants to lane follow along the road travelling eastbound. There is a hero vehicle on the driveway. It wants to turn left onto the road travelling eastbound. When the ego is ego\_distance\_behind\_conflict\_point\_m behind the conflict point and the hero vehicle is hero\_distance\_behind\_stop\_line\_m behind the stop line, the hero vehicle begins to accelerate to complete the turn before the ego passes the conflict point.
    \item Ego is on the road and it is initial\_ego\_distance\_behind\_intersection\_m behind the intersection. It wants to lane follow along the road travelling eastbound. There is a hero vehicle on the driveway. It wants to turn left onto the road travelling eastbound. When the ego is ego\_distance\_behind\_conflict\_point\_m behind the conflict point and the hero vehicle is hero\_distance\_behind\_stop\_line\_m behind the stop line, the hero vehicle begins to decelerate to a complete stop just before the stop line.  Then, the hero vehicle begins to accelerate to complete the turn before the ego passes the conflict point.
    \item Ego is on the road and it is initial\_ego\_distance\_behind\_intersection\_m behind the intersection. It wants to turn left into the driveway. There is a hero vehicle on the road. It wants to lane follow along the road travelling westbound. When the ego is ego\_distance\_behind\_conflict\_point\_m behind the conflict point and the hero vehicle is hero\_distance\_behind\_conflict\_point\_m behind the conflict point, the hero vehicle begins to decelerate to a complete stop at end\_hero\_distance\_behind\_conflict\_point\_m behind the conflict point.
    \item Ego is on the road and it is initial\_ego\_distance\_behind\_intersection\_m behind the intersection. It wants to turn left into the driveway. There is a hero vehicle on the road. It wants to lane follow along the road travelling westbound. When the ego is ego\_distance\_behind\_conflict\_point\_m behind the conflict point and the hero vehicle is distance\_behind\_conflict\_point\_m behind the conflict point, the hero vehicle begins to accelerate to pass the conflict point before the ego begins to turn.
    \item Ego is on the road and it is initial\_ego\_distance\_behind\_intersection\_m behind the intersection. It wants to turn left into the driveway. There is a hero vehicle on the road. It wants to lane follow along the road travelling westbound. When the ego is ego\_distance\_behind\_conflict\_point\_m behind the conflict point and the hero vehicle is distance\_behind\_conflict\_point\_m behind the conflict point, the hero vehicle begins to decelerate to a complete stop at end\_hero\_distance\_behind\_conflict\_point\_m behind the conflict point. Then, the hero vehicle begins to accelerate to pass the conflict point before the ego begins to turn.
    \item Ego is on the road and it is initial\_ego\_distance\_behind\_intersection\_m behind the intersection. It wants to turn left into the driveway. There is a hero vehicle on the road. It wants to turn right into the driveway. When the ego is ego\_distance\_behind\_conflict\_point\_m behind the conflict point and the hero vehicle is hero\_distance\_behind\_conflict\_point\_m behind the conflict point, the hero vehicle begins to decelerate to a complete stop at end\_hero\_distance\_behind\_conflict\_point\_m behind the conflict point.
    \item Ego is on the road and it is initial\_ego\_distance\_behind\_intersection\_m behind the intersection. It wants to turn left into the driveway. There is a hero vehicle on the road. It wants to turn right into the driveway. When the ego is ego\_distance\_behind\_conflict\_point\_m behind the conflict point and the hero vehicle is distance\_behind\_conflict\_point\_m behind the conflict point, the hero vehicle begins to accelerate to pass the conflict point before the ego begins to turn.
    \item Ego is on the road and it is initial\_ego\_distance\_behind\_intersection\_m behind the intersection. It wants to turn left into the driveway. There is a hero vehicle on the road. It wants to turn right into the driveway. When the ego is ego\_distance\_behind\_conflict\_point\_m behind the conflict point and the hero vehicle is distance\_behind\_conflict\_point\_m behind the conflict point, the hero vehicle begins to decelerate to a complete stop at end\_hero\_distance\_behind\_conflict\_point\_m behind the conflict point. Then, the hero vehicle begins to accelerate to pass the conflict point before the ego begins to turn.
    \item Ego is on the road and it is initial\_ego\_distance\_behind\_intersection\_m behind the intersection. It wants to turn left into the driveway. There is a hero vehicle on the driveway. It wants to turn left onto the road travelling eastbound. When the ego is ego\_distance\_behind\_conflict\_point\_m behind the conflict point and the hero vehicle is hero\_distance\_behind\_stop\_line\_m behind the stop line, the hero vehicle begins to decelerate to a complete stop just before the stop line.
    \item Ego is on the road and it is initial\_ego\_distance\_behind\_intersection\_m behind the intersection. It wants to turn left into the driveway. There is a hero vehicle on the driveway. It wants to turn left onto the road travelling eastbound. When the ego is ego\_distance\_behind\_conflict\_point\_m behind the conflict point  and the hero vehicle is hero\_distance\_behind\_stop\_line\_m behind the stop line, the hero vehicle begins to accelerate to pass the conflict point before the ego passes the conflict point.
    \item Ego is on the road and it is initial\_ego\_distance\_behind\_intersection\_m behind the intersection. It wants to turn left into the driveway. There is a hero vehicle on the driveway. It wants to turn left onto the road travelling eastbound. When the ego is ego\_distance\_behind\_conflict\_point\_m behind the conflict point  and the hero vehicle is hero\_distance\_behind\_stop\_line\_m behind the stop line, the hero vehicle begins to decelerate to a complete stop just before the stop line. Then, the hero vehicle begins to accelerate to pass the conflict point before the ego passes the conflict point.
    \item Ego is on the driveway exit and it is initial\_ego\_tta\_behind\_stop\_line\_s from the stop line. It wants to turn right onto the road travelling westbound. There is a hero vehicle on the driveway exit and it is initial\_hero\_tta\_behind\_stop\_line\_s from the stop line. It wants to turn right onto the road travelling westbound. When the ego is ego\_distance\_behind\_hero\_m behind the hero vehicle, the hero vehicle begins to decelerate to a complete stop at a rate of hero\_deceleration\_mpss.
    \item Ego is on the driveway exit and it is initial\_ego\_tta\_behind\_stop\_line\_s from the stop line. It wants to turn right onto the road travelling westbound. There is a hero vehicle on the road. It wants to lane follow along the road travelling westbound. When the ego is ego\_tta\_behind\_stop\_line\_s from the stop line and the hero vehicle is hero\_tta\_behind\_conflict\_point\_s from the conflict point, the hero vehicle begins to decelerate to a complete stop at end\_hero\_distance\_behind\_conflict\_point\_m behind the conflict point.
    \item Ego is on the driveway exit and it is initial\_ego\_tta\_behind\_stop\_line\_s from the stop line. It wants to turn right onto the road travelling westbound. There is a hero vehicle on the road. It wants to lane follow along the road travelling westbound. When the ego is ego\_tta\_behind\_stop\_line\_s from the stop line and the hero vehicle is tta\_behind\_conflict\_point\_s from the conflict point, the hero vehicle begins to accelerate to pass the conflict point before the ego begins to turn.
    \item Ego is on the driveway exit and it is initial\_ego\_tta\_behind\_stop\_line\_s from the stop line. It wants to turn right onto the road travelling westbound. There is a hero vehicle on the road. It wants to lane follow along the road travelling westbound. When the ego is ego\_tta\_behind\_stop\_line\_s from the stop line and the hero vehicle is tta\_behind\_conflict\_point\_s from the conflict point, the hero vehicle begins to decelerate to a complete stop at end\_hero\_distance\_behind\_conflict\_point\_m behind the conflict point. Then, the hero vehicle begins to accelerate to pass the conflict point before the ego begins to turn.
    \item Ego is on the driveway exit and it is initial\_ego\_tta\_behind\_stop\_line\_s from the stop line. It wants to turn left onto the road travelling eastbound. There is a hero vehicle on the driveway exit and it is initial\_hero\_tta\_behind\_stop\_line\_s from the stop line. It wants to turn left onto the road travelling eastbound. When the ego is ego\_distance\_behind\_hero\_m behind the hero vehicle, the hero vehicle begins to decelerate to a complete stop at a rate of hero\_deceleration\_mpss.
    \item Ego is on the driveway exit and it is initial\_ego\_tta\_behind\_stop\_line\_s from the stop line. It wants to turn left onto the road travelling eastbound. There is a hero vehicle on the road. It wants to lane follow along the road travelling westbound. When the ego is ego\_tta\_behind\_stop\_line\_s from the stop line and the hero vehicle is hero\_tta\_behind\_conflict\_point\_s from the conflict point, the hero vehicle begins to decelerate to a complete stop at end\_hero\_distance\_behind\_conflict\_point\_m behind the conflict point.
    \item Ego is on the driveway exit and it is initial\_ego\_tta\_behind\_stop\_line\_s from the stop line. It wants to turn left onto the road travelling eastbound. There is a hero vehicle on the road. It wants to lane follow along the road travelling westbound. When the ego is ego\_tta\_behind\_stop\_line\_s from the stop line and the hero vehicle is tta\_behind\_conflict\_point\_s from the conflict point, the hero vehicle begins to accelerate to pass the conflict point before the ego begins to turn.
    \item Ego is on the driveway exit and it is initial\_ego\_tta\_behind\_stop\_line\_s from the stop line. It wants to turn left onto the road travelling eastbound. There is a hero vehicle on the road. It wants to lane follow along the road travelling westbound. When the ego is ego\_tta\_behind\_stop\_line\_s from the stop line and the hero vehicle is tta\_behind\_conflict\_point\_s from the conflict point, the hero vehicle begins to decelerate to a complete stop at end\_hero\_distance\_behind\_conflict\_point\_m behind the conflict point. Then, the hero vehicle begins to accelerate to pass the conflict point before the ego begins to turn.
    \item Ego is on the driveway exit and it is initial\_ego\_tta\_behind\_stop\_line\_s from the stop line. It wants to turn left onto the road travelling eastbound. There is a hero vehicle on the road. It wants to lane follow along the road travelling eastbound. When the ego is ego\_tta\_behind\_stop\_line\_s from the stop line and the hero vehicle is hero\_tta\_behind\_conflict\_point\_s from the conflict point, the hero vehicle begins to decelerate to a complete stop at end\_hero\_distance\_behind\_conflict\_point\_m behind the conflict point.
    \item Ego is on the driveway exit and it is initial\_ego\_tta\_behind\_stop\_line\_s from the stop line. It wants to turn left onto the road travelling eastbound. There is a hero vehicle on the road. It wants to lane follow along the road travelling eastbound. When the ego is ego\_tta\_behind\_stop\_line\_s from the stop line and the hero vehicle is tta\_behind\_conflict\_point\_s from the conflict point, the hero vehicle begins to accelerate to pass the conflict point before the ego begins to turn.
    \item Ego is on the driveway exit and it is initial\_ego\_tta\_behind\_stop\_line\_s from the stop line. It wants to turn left onto the road travelling eastbound. There is a hero vehicle on the road. It wants to lane follow along the road travelling eastbound. When the ego is ego\_tta\_behind\_stop\_line\_s from the stop line and the hero vehicle is tta\_behind\_conflict\_point\_s from the conflict point, the hero vehicle begins to decelerate to a complete stop at end\_hero\_distance\_behind\_conflict\_point\_m behind the conflict point. Then, the hero vehicle begins to accelerate to pass the conflict point before the ego begins to turn.
    \item Ego is on the driveway exit and it is initial\_ego\_tta\_behind\_stop\_line\_s from the stop line. It wants to turn left onto the road travelling eastbound. There is a hero vehicle on the road. It wants to turn left into the driveway. When the ego is ego\_tta\_behind\_stop\_line\_s from the stop line and the hero vehicle is hero\_tta\_behind\_conflict\_point\_s from the conflict point, the hero vehicle begins to decelerate to a complete stop at end\_hero\_distance\_behind\_conflict\_point\_m behind the conflict point.
    \item Ego is on the driveway exit and it is initial\_ego\_tta\_behind\_stop\_line\_s from the stop line. It wants to turn left onto the road travelling eastbound. There is a hero vehicle on the road. It wants to turn left into the driveway. When the ego is ego\_tta\_behind\_stop\_line\_s from the stop line and the hero vehicle is tta\_behind\_conflict\_point\_s from the conflict point, the hero vehicle begins to accelerate to pass the conflict point before the ego begins to turn.
    \item Ego is on the driveway exit and it is initial\_ego\_tta\_behind\_stop\_line\_s from the stop line. It wants to turn left onto the road travelling eastbound. There is a hero vehicle on the road. It wants to turn left into the driveway. When the ego is ego\_tta\_behind\_stop\_line\_s from the stop line and the hero vehicle is tta\_behind\_conflict\_point\_s from the conflict point, the hero vehicle begins to decelerate to a complete stop at end\_hero\_distance\_behind\_conflict\_point\_m behind the conflict point. Then, the hero vehicle begins to accelerate to pass the conflict point before the ego begins to turn.
    \item Ego is on the road and it is initial\_ego\_tta\_behind\_intersection\_s from the intersection. It wants to lane follow along the road travelling westbound. There is a hero vehicle on the road. It wants to lane follow along the road travelling westbound. When the ego is ego\_distance\_behind\_hero\_m behind the hero vehicle, the hero vehicle begins to decelerate to a complete stop at a rate of hero\_deceleration\_mpss.
    \item Ego is on the road and it is initial\_ego\_tta\_behind\_intersection\_s from the intersection. It wants to lane follow along the road travelling westbound. There is a hero vehicle on the road. It wants to turn left into the driveway. When the ego is ego\_tta\_behind\_conflict\_point\_s from the conflict point and the hero vehicle is hero\_tta\_behind\_conflict\_point\_s from the conflict point, the hero vehicle begins to decelerate to a complete stop at end\_hero\_distance\_behind\_conflict\_point\_m behind the conflict point.
    \item Ego is on the road and it is initial\_ego\_tta\_behind\_intersection\_s from the intersection. It wants to lane follow along the road travelling westbound. There is a hero vehicle on the road. It wants to turn left into the driveway. When the ego is ego\_tta\_behind\_conflict\_point\_s from the conflict point and the hero vehicle is hero\_tta\_behind\_conflict\_point\_s from the conflict point, the hero vehicle begins to accelerate to pass the conflict point before the ego begins to turn.
    \item Ego is on the road and it is initial\_ego\_tta\_behind\_intersection\_s from the intersection. It wants to lane follow along the road travelling westbound. There is a hero vehicle on the road. It wants to turn left into the driveway. When the ego is ego\_tta\_behind\_conflict\_point\_s from the conflict point and the hero vehicle is hero\_tta\_behind\_conflict\_point\_s from the conflict point, the hero vehicle begins to decelerate to a complete stop at end\_hero\_distance\_behind\_conflict\_point\_m behind the conflict point. Then, the hero vehicle begins to accelerate to pass the conflict point before the ego passes the conflict point.
    \item Ego is on the road and it is initial\_ego\_tta\_behind\_intersection\_s from the intersection. It wants to lane follow along the road travelling westbound. There is a hero vehicle on the driveway. It wants to turn right onto the road travelling westbound. When the ego is ego\_tta\_behind\_conflict\_point\_s from the conflict point and the hero vehicle is hero\_tta\_behind\_stop\_line\_s from the stop line, the hero vehicle begins to decelerate to a complete stop just before the stop line.
    \item Ego is on the road and it is initial\_ego\_tta\_behind\_intersection\_s from the intersection. It wants to lane follow along the road travelling westbound. There is a hero vehicle on the driveway. It wants to turn right onto the road travelling westbound. When the ego is ego\_tta\_behind\_conflict\_point\_s from the conflict point and the hero vehicle is hero\_tta\_behind\_stop\_line\_s from the stop line, the hero vehicle begins to accelerate to complete the turn before the ego passes the conflict point.
    \item Ego is on the road and it is initial\_ego\_tta\_behind\_intersection\_s from the intersection. It wants to lane follow along the road travelling westbound. There is a hero vehicle on the driveway. It wants to turn right onto the road travelling westbound. When the ego is ego\_tta\_behind\_conflict\_point\_s from the conflict point and the hero vehicle is hero\_tta\_behind\_stop\_line\_s from the stop line, the hero vehicle begins to decelerate to a complete stop just before the stop line. Then, the hero vehicle begins to accelerate to complete the turn before the ego passes the conflict point.
    \item Ego is on the road and it is initial\_ego\_tta\_behind\_intersection\_s from the intersection. It wants to lane follow along the road travelling westbound. There is a hero vehicle on the driveway. It wants to turn left onto the road travelling eastbound. When the ego is ego\_tta\_behind\_conflict\_point\_s from the conflict point and the hero vehicle is hero\_tta\_behind\_stop\_line\_s from the stop line, the hero vehicle begins to decelerate to a complete stop just before the stop line.
    \item Ego is on the road and it is initial\_ego\_tta\_behind\_intersection\_s from the intersection. It wants to lane follow along the road travelling westbound. There is a hero vehicle on the driveway. It wants to turn left onto the road travelling eastbound. When the ego is ego\_tta\_behind\_conflict\_point\_s from the conflict point and the hero vehicle is hero\_tta\_behind\_stop\_line\_s from the stop line, the hero vehicle begins to accelerate to complete the turn before the ego passes the conflict point.
    \item Ego is on the road and it is initial\_ego\_tta\_behind\_intersection\_s from the intersection. It wants to lane follow along the road travelling westbound. There is a hero vehicle on the driveway. It wants to turn left onto the road travelling eastbound. When the ego is ego\_tta\_behind\_conflict\_point\_s from the conflict point and the hero vehicle is hero\_tta\_behind\_stop\_line\_s from the stop line, the hero vehicle begins to decelerate to a complete stop just before the stop line. Then, the hero vehicle begins to accelerate to complete the turn before the ego passes the conflict point.
    \item Ego is on the road and it is initial\_ego\_tta\_behind\_intersection\_s from the intersection. It wants to lane follow along the road travelling eastbound. There is a hero vehicle on the road. It wants to lane follow along the road travelling eastbound. When the ego is ego\_distance\_behind\_hero\_m behind the hero vehicle, the hero vehicle begins to decelerate to a complete stop at a rate of hero\_deceleration\_mpss.
    \item Ego is on the road and it is initial\_ego\_tta\_behind\_intersection\_s from the intersection. It wants to lane follow along the road travelling eastbound. There is a hero vehicle on the driveway. It wants to turn left onto the road travelling eastbound. When the ego is ego\_tta\_behind\_conflict\_point\_s from the conflict point and the hero vehicle is hero\_tta\_behind\_stop\_line\_s from the stop line, the hero vehicle begins to decelerate to a complete stop just before the stop line.
    \item Ego is on the road and it is initial\_ego\_tta\_behind\_intersection\_s from the intersection. It wants to lane follow along the road travelling eastbound. There is a hero vehicle on the driveway. It wants to turn left onto the road travelling eastbound. When the ego is ego\_tta\_behind\_conflict\_point\_s from the conflict point and the hero vehicle is hero\_tta\_behind\_stop\_line\_s from the stop line, the hero vehicle begins to accelerate to complete the turn before the ego passes the conflict point.
    \item Ego is on the road and it is initial\_ego\_tta\_behind\_intersection\_s from the intersection. It wants to lane follow along the road travelling eastbound. There is a hero vehicle on the driveway. It wants to turn left onto the road travelling eastbound. When the ego is ego\_tta\_behind\_conflict\_point\_s from the conflict point and the hero vehicle is hero\_tta\_behind\_stop\_line\_s from the stop line, the hero vehicle begins to decelerate to a complete stop just before the stop line. Then, the hero vehicle begins to accelerate to complete the turn before the ego passes the conflict point.
    \item Ego is on the road and it is initial\_ego\_tta\_behind\_intersection\_s from the intersection. It wants to turn right into the driveway. There is a hero vehicle on the road. It wants to turn left into the driveway. When the ego is ego\_tta\_behind\_conflict\_point\_s from the conflict point and the hero vehicle is hero\_tta\_behind\_conflict\_point\_s from the conflict point, the hero vehicle begins to decelerate to a complete stop at end\_hero\_distance\_behind\_conflict\_point\_m behind the conflict point.
    \item Ego is on the road and it is initial\_ego\_tta\_behind\_intersection\_s from the intersection. It wants to lane follow along the road travelling westbound. There is a hero vehicle on the road. It wants to turn left into the driveway. When the ego is ego\_tta\_behind\_conflict\_point\_s from the conflict point and the hero vehicle is hero\_tta\_behind\_conflict\_point\_s from the conflict point, the hero vehicle begins to accelerate to pass the conflict point before the ego passes the conflict point.
    \item Ego is on the road and it is initial\_ego\_tta\_behind\_intersection\_s from the intersection. It wants to lane follow along the road travelling westbound. There is a hero vehicle on the road. It wants to turn left into the driveway. When the ego is ego\_tta\_behind\_conflict\_point\_s from the conflict point and the hero vehicle is tta\_behind\_conflict\_point\_s from the conflict point, the hero vehicle begins to decelerate to a complete stop at end\_hero\_distance\_behind\_conflict\_point\_m behind the conflict point. Then, the hero vehicle begins to accelerate to pass the conflict point before the ego passes the conflict point.
    \item Ego is on the road and it is initial\_ego\_tta\_behind\_intersection\_s from the intersection. It wants to turn left into the driveway. There is a hero vehicle on the road. It wants to lane follow along the road travelling westbound. When the ego is ego\_tta\_behind\_conflict\_point\_s from the conflict point and the hero vehicle is hero\_tta\_behind\_conflict\_point\_s from the conflict point, the hero vehicle begins to decelerate to a complete stop at end\_hero\_distance\_behind\_conflict\_point\_m behind the conflict point.
    \item Ego is on the road and it is initial\_ego\_tta\_behind\_intersection\_s from the intersection. It wants to turn left into the driveway. There is a hero vehicle on the road. It wants to lane follow along the road travelling westbound. When the ego is ego\_tta\_behind\_conflict\_point\_s from the conflict point and the hero vehicle is tta\_behind\_conflict\_point\_s from the conflict point, the hero vehicle begins to accelerate to pass the conflict point before the ego begins to turn.
    \item Ego is on the road and it is initial\_ego\_tta\_behind\_intersection\_s from the intersection. It wants to turn left into the driveway. There is a hero vehicle on the road. It wants to lane follow along the road travelling westbound. When the ego is ego\_tta\_behind\_conflict\_point\_s from the conflict point and the hero vehicle is tta\_behind\_conflict\_point\_s from the conflict point, the hero vehicle begins to decelerate to a complete stop at end\_hero\_distance\_behind\_conflict\_point\_m behind the conflict point. Then, the hero vehicle begins to accelerate to pass the conflict point before the ego begins to turn.
    \item Ego is on the road and it is initial\_ego\_tta\_behind\_intersection\_s from the intersection. It wants to turn left into the driveway. There is a hero vehicle on the road. It wants to turn right into the driveway. When the ego is ego\_tta\_behind\_conflict\_point\_s from the conflict point and the hero vehicle is hero\_tta\_behind\_conflict\_point\_s from the conflict point, the hero vehicle begins to decelerate to a complete stop at end\_hero\_distance\_behind\_conflict\_point\_m behind the conflict point.
    \item Ego is on the road and it is initial\_ego\_tta\_behind\_intersection\_s from the intersection. It wants to turn left into the driveway. There is a hero vehicle on the road. It wants to turn right into the driveway. When the ego is ego\_tta\_behind\_conflict\_point\_s from the conflict point and the hero vehicle is tta\_behind\_conflict\_point\_s from the conflict point, the hero vehicle begins to accelerate to pass the conflict point before the ego begins to turn.
    \item Ego is on the road and it is initial\_ego\_tta\_behind\_intersection\_s from the intersection. It wants to turn left into the driveway. There is a hero vehicle on the road. It wants to turn right into the driveway. When the ego is ego\_tta\_behind\_conflict\_point\_s from the conflict point and the hero vehicle is tta\_behind\_conflict\_point\_s from the conflict point, the hero vehicle begins to decelerate to a complete stop at end\_hero\_distance\_behind\_conflict\_point\_m behind the conflict point. Then, the hero vehicle begins to accelerate to pass the conflict point before the ego begins to turn.
    \item Ego is on the road and it is initial\_ego\_tta\_behind\_intersection\_s from the intersection. It wants to turn left into the driveway. There is a hero vehicle on the driveway. It wants to turn left onto the road travelling eastbound. When the ego is ego\_tta\_behind\_conflict\_point\_s from the conflict point and the hero vehicle is hero\_tta\_behind\_stop\_line\_s from the stop line, the hero vehicle begins to decelerate to a complete stop just before the stop line.
    \item Ego is on the road and it is initial\_ego\_tta\_behind\_intersection\_s from the intersection. It wants to turn left into the driveway. There is a hero vehicle on the driveway. It wants to turn left onto the road travelling eastbound. When the ego is ego\_tta\_behind\_conflict\_point\_s from the conflict point and the hero vehicle is hero\_tta\_behind\_stop\_line\_s from the stop line, the hero vehicle begins to accelerate to pass the conflict point before the ego passes the conflict point.
    \item Ego is on the road and it is initial\_ego\_tta\_behind\_intersection\_s from the intersection. It wants to turn left into the driveway. There is a hero vehicle on the driveway. It wants to turn left onto the road travelling eastbound. When the ego is ego\_tta\_behind\_conflict\_point\_s from the conflict point and the hero vehicle is hero\_tta\_behind\_stop\_line\_s from the stop line, the hero vehicle begins to decelerate to a complete stop just before the stop line. Then, the hero vehicle begins to accelerate to pass the conflict point before the ego passes the conflict point.
\end{enumerate}

\section{Open- vs. Closed-Loop Benchmark Scenario Descriptions}
\label{sec:open-vs-closed-loop-desc}
\begin{enumerate}
  \item \textbf{lead\_actor-turn\_into\_driveway-empty\_world}: Ego is lane-following straight along the road at \{ego\_initial\_speed\_mps\}m/s. There is a driveway 100m in front of the ego that merges into the ego's lane. There is a lead vehicle that is initially driving at \{initial\_speed\_mps\}m/s. When the lead vehicle is {distance\_ahead\_of\_ego\_m}m ahead of the ego, it turns right into the driveway and gradually comes to a stop.
  \item \textbf{oncoming\_actor-turn\_into\_driveway-empty\_world}: Ego is lane-following straight along the road at \{ego\_initial\_speed\_mps\}m/s. There is a driveway 100m in front of the ego that merges into the ego's lane. There is a car in the oncoming lane driving straight. When the time-to-collision between the ego and the car is \{ttc\_s\}s, the car should arrive at the mouth of the driveway and turn left across the ego's path and into the driveway.
  \item \textbf{actor\_on\_driveway-turn\_into\_path-empty\_world}: Ego is lane-following straight along the road at \{ego\_initial\_speed\_mps\}m/s. There is a driveway 100m in front of the ego that merges into the ego's lane. There is a car on the driveway that turns right into the ego's lane when the ego is about \{tta\_s\}s away from the driveway. After turning right, the car starts lane-following straight at \{target\_speed\_mps\}m/s.
  \item \textbf{actor\_on\_driveway-turn\_across\_path-empty\_world}: Ego is lane-following straight along the road at \{ego\_initial\_speed\_mps\}m/s. There is a driveway 100m in front of the ego that merges into the ego's lane. There is a car on the driveway that turns left across the ego's path onto the ego's oncoming lane when the ego is about \{tta\_s\}s away from the driveway.
\end{enumerate}

\section{Example Descriptions with Pseudocode Constraints}
\label{sec:pseudocode}

\subsection{Lead actor turn into driveway empty world}

\begin{verbatim}
Ego is lane-following straight along the road at {ego_initial_speed_mps}m/s. 
There is a driveway 250m in front of the ego that merges into the ego's lane. 
There is a lead vehicle that is initially driving at {initial_speed_mps}m/s. 
When the lead vehicle is {distance_ahead_of_ego_m}m ahead of the ego, 
it turns right into the driveway and gradually comes to a stop.
\end{verbatim}

\begin{verbatim}
Actor 0: 
- W
- [t0, go, t1, go, t2]
Actor 1: 
- W, N
- [t0, go, t1, dec, t2]

Constraints:
A0v(t0) == ego_initial_speed_mps
A0x(t0) == turn - 100
A1v(t0) == initial_speed_mps
A1x(t1) - A0x(t1) == distance_ahead_of_ego_m
A1x(t1) == turn
A1v(t2) == 0
A0(t1) == A1(t1)
\end{verbatim}

\subsection{Right turn driveway actor hesitate and go empty world}

\begin{verbatim}
Ego is on the driveway. It is driving at {ego_initial_speed_mps}m/s. 
It wants to make a right turn onto the road. 
There is a car in the lane that the ego wants to turn into. 
The car is initially driving at {initial_speed_mps}m/s. 
It decelerates at a rate of {deceleration_mpss}m/s^2 to a complete stop,
as if to yield, exactly {distance_to_driveway_m}m from the driveway.
It then re-accelerates back to {initial_speed_mps}m/s.
\end{verbatim}

\begin{verbatim}
Actor 0:
- S, W
- [t0, go, t1, go, t2]
Actor 1:
- W
- [t0, dec, t1, stop, t2, acc, t3]

Constraints:
A0v(t0) == ego_initial_speed_mps
A1v(t0) == initial_speed_mps
A1a(dec) == deceleration_mpss
A1v(t1) == 0
A1x(t1) == turn - distance_to_driveway_m
A1v(stop) == 0
A1a(stop) == 0
A0x(t1) == turn
A0(t1) == A1(t1)
A1(t2) > A0(t1)
A1v(t3) == initial_speed_mps
\end{verbatim}

\section{Full-Length Prompts}

\noindent
\begin{minipage}{\linewidth}
    \scalebox{0.95}{
  \begin{tcolorbox}[title={\textbf{\scriptsize Natural Language to Constraint Generation Prompt}}, enhanced, breakable=false, colback=white, colframe=black, fontupper=\scriptsize]
    You convert descriptions of surface street driving scenarios into actor trajectories and pseudocode constraints.
    The scenario consists of exactly two actors, one ego vehicle (id = 0) and one other actor (id = 1).

    \# Actor trajectories

    An actor trajectory consists of a spatial route and a piecewise polynomial decomposition.
    Spatial route: a list of cardinal directions that the actor is headed in.
    For example, if an actor is headed from West to East, the route should be "E".
    Piecewise polynomial decomposition: the representation of the trajectory. It is defined by a sequence of knots interleaved with pieces.
    A piece can be "go", "acc" (for accelerate), "dec" (for decelerate), or "stop". A knot is named "t{idx}" where idx is the index starting at 0.

    \#\# Map

    The map is a T-intersection with a driveway to the North and the two-way road running East-West and West-East.
    If an actor is on the driveway exit, it is headed South. 
    
    If an actor ends up on the driveway entrance, it is headed North. 
    If an actor is on the road, it is either headed West or East. 
    Infer the direction based on whether the actor wants to turn left or right into the driveway.

    \# Pseudocode constraints

    To refer to the acceleration of an actor with id `i` at knot `t\{j\}`, use `A\{i\}a("t\{j\}")`.
    To refer to the velocity of an actor with id `i` at knot `t\{j\}`, use `A\{i\}v("t\{j\}")`.
    To refer to the position of an actor with id `i` at knot `t\{j\}`, use `A\{i\}x("t\{j\}")`.
    To refer to the duration of an actor with id `i` up to knot `t\{j\}`, use `A\{i\}("t\{j\}")`.

    By default, named, symbolic time points corresponding to different actors have no relation to each other. 
    It is necessary to manually relate named, symbolic time points of different actors to each other if they refer to the same interval.

    The scenario has parameters indicated by their snake case names, which always end in a unit (\_m for meters or \_s for seconds).
    Note that deceleration is signed to be negative.

    There are special constants `stop\_line`, `conflict\_point`, `turn\_start`, and `turn\_end`.

    Notes:

    - The intersection should be mapped to the conflict point.

    - Translate the words "just before" to the location minus 5.0m; i.e. "just before the stop line" should be mapped to `stop\_line - 5.0`.

    **Important**: Make sure the LHS of each constraint (in)equality is an expression involving an actor state. 
    You may need to swap the order of the LHS and RHS of a constraint in order to ensure this.
  \end{tcolorbox}}
  \vspace{5pt}
\end{minipage}
\noindent
\begin{minipage}{\linewidth}
    \scalebox{0.95}{
  \begin{tcolorbox}[title={\textbf{\scriptsize Constraint to Scenario Class Generation Prompt}}, enhanced, breakable=false, colback=white, colframe=black, fontupper=\scriptsize]
    You translate actor trajectories and pseudocode constraints into a Python class.

    The Python class should inherit from the Scenario class and be named the scenario name converted to camel case and with "Scenario" appended.

    A Scenario class defines .reset() and .constraints() methods.

    The reset method takes as arguments the scenario parameters. These are variables indicated by their snake case names and which always end in a unit ("\_s" or "\_m").
    The reset method instantiates the actors in the scenario by defining their routes and trajectories.

    The map\_adapter provides a function `create\_route` which takes in any number of strings where each string is a character representing a cardinal direction.
    These should be the route segments given in the actor trajectory.

    To get the turn points, check whether `segments\_ego` and `segments\_oth` match at the end.
    If their last characters do not match, use `map\_adapter.get\_intersection(route\_ego, route\_oth)`.
    If their last characters do match, use `map\_adapter.get\_segment\_intersection(segments\_ego, segments\_oth)`.
    Both functions return a point, the fraction for the first segments, and the fraction for the second segments.

    The pseudocode trajectories consist of interleaved knots and pieces.
    When converting, give each piece a globally unique name (e.g., append it with a unique numerical identifier).
    Give the Pieces Orders corresponding to the following mapping from their base name to their Order:

    - go \texttt{-->} Order.VELOCITY

    - acc \texttt{-->} Order.ACCELERATION

    - dec \texttt{-->} Order.ACCELERATION

    - stop \texttt{-->} Order.POSITION

    The constraints method returns a list of Z3 constraints based on the pseudocode constraints.

    In the pseudocode constraints, `A\{id\}\{order\}(t)` refers to the state of order `order` of the actor with id `id` at time `t`.
    Note that in the pseudocode constraints, `A\{id\}(t)` (without a 'x', 'v', or 'a' after \{id\}) represents a symbolic duration up to the knot `t`.

    To get the closed-form, algebraic expression of the actor state at some symbolic time t, use the Python code: `actors[id: int].get(order: Order)(t: str)`.
    To get the closed-form, algebraic expression of the actor state at some concrete time t, use the Python code: `actors[id: int].calc(order: Order)(t: float)`.
    To get the duration of a trajectory up to a given knot t, use the Python code: `actors[id: int].trajectory.\_duration(t)`.

    In the pseudocode constraints, there are special constants `stop\_line`, `conflict\_point`, `turn\_start`, and `turn\_end`.
    These should be mapped as follows:
    - stop\_line \texttt{-->} `distance\_func("turn\_start")`

    - conflict\_point \texttt{-->} `distance\_func("lane\_turn")`

    - turn\_start \texttt{-->} `distance\_func("turn\_start")`

    - turn\_end \texttt{-->} `distance\_func("turn\_end")`
  \end{tcolorbox}}
  \vspace{5pt}
\end{minipage}
\noindent
\begin{minipage}{\linewidth}
    \scalebox{0.95}{
  \begin{tcolorbox}[title={\textbf{\scriptsize Open Loop to Closed Loop Scenario Class Generation Prompt}}, enhanced, breakable=false, colback=white, colframe=black, fontupper=\scriptsize]
    Modify the given Scenario class as follows:

    - add self.search = True in the .reset() method

    - cast the MapAdapter to the type Lancaster\_T\_Intersection

    - for the ego actor, relax the order of each piece to one higher order, or leave it as the same if the order is the maximum (i.e. Order.ACCELERATION)

    \texttt{    } - **IMPORTANT**: make sure that all pieces for both actors continue to have globally unique names even after lifting the ego pieces up to order acceleration.

    - fix the fraction calculations by replacing the code with a single call to map\_adapter.get\_intersection()

    - if segments\_ego starts with "S" (meaning the ego starts on the driveway), add the following constraints:
   
    `ego.get(Order.POSITION)("t0") \texttt{<=} distance\_ego("lane\_start", "lane\_start") + 50`
   
    `ego.get(Order.POSITION)("t0") \texttt{>=} distance\_ego("lane\_start", "lane\_start") + 40`
   
    and place them inside a conditional so that they only apply when the
    orchestration mode is OPEN\_LOOP or CLOSED\_LOOP\_INITIALIZATION

    \texttt{    }  - otherwise, add the following constraints:

    `ego.get(Order.POSITION)("t0") \texttt{<=} distance\_ego("lane\_turn") - 50)`
    and place them inside a conditional so that they only apply when the orchestration mode is OPEN\_LOOP or CLOSED\_LOOP\_INITIALIZATION

    - for each constraint that sets the initial state (position, velocity, or acceleration) of an actor, place it inside a conditional so that it only applies when the orchestration mode is OPEN\_LOOP or CLOSED\_LOOP\_INITIALIZATION

    - identify the reactive constraints. These will be constraints that has ego.get(Order.POSITION) or oth.get(Order.POSITION) at an intermediate time point (e.g. "t1" or in rare cases, "t2" if the actor in the scenario has a "t3" in its trajectory formulation) on the left hand side.

    If the orchestration mode is OPEN\_LOOP or CLOSED\_LOOP\_INITIALIZATION, leave it unchanged.

    If the orchestration mode is CLOSED\_LOOP\_ROLLOUT, then assert `world\_state`, set the `buffer` amount to be `world\_state.value`, and weaken the inequality by `buffer` on either side.

    Output the complete, modified Scenario class after making all the requested modifications. Do not output anything else.
  \end{tcolorbox}}
\end{minipage}